%% file: main.tex
\documentclass[11pt]{article}

\usepackage[preprint]{acl}
\usepackage{placeins}

\usepackage{times}
\usepackage{enumitem}
\usepackage{latexsym}
\usepackage{pifont}
\usepackage{xcolor}
\usepackage{booktabs} 
\usepackage{multirow} 
\usepackage[table]{xcolor}
\usepackage{colortbl}
\usepackage{array}
\usepackage{soul}
\usepackage{url}
\usepackage{amsmath} 
\usepackage{graphicx} 
\usepackage{colortbl} 
\usepackage{makecell} 
\usepackage{amssymb}

\usepackage{subcaption}
\usepackage{tcolorbox} 
\usepackage{fontawesome5}
\newcommand{\pro}{\textcolor{green!50!black}{\faThumbsUp}}
\newcommand{\con}{\textcolor{red!70!black}{\faThumbsDown}}

\usepackage[T1]{fontenc}

\usepackage[utf8]{inputenc}

\usepackage{microtype}

\usepackage{inconsolata}
\usepackage{xcolor}
\usepackage[table]{xcolor}
\newcounter{todocounter}
\usepackage{bm}

\usepackage{soul}      
\usepackage{xcolor} 
\definecolor{bestlightblue}{RGB}{210,225,240}   
\definecolor{bestcoral}{RGB}{240,220,210}        
\definecolor{bestlavender}{RGB}{225,220,238}

\newcommand{\up}{\bm{\uparrow}}
\newcommand{\down}{\bm{\downarrow}}

\usepackage{fontawesome5}

\title{Disagreeing Rationales: Rethinking Classification and Explainability Evaluation in Hate Speech Detection}

\author{
\textbf{Benedetta Muscato\textsuperscript{\faPizzaSlice\kern1pt\faGopuram}} \quad
\textbf{Beiduo Chen\textsuperscript{\faMountain\kern1pt\faRobot}} \quad
\textbf{Gizem Gezici\textsuperscript{\faPizzaSlice}} \\
\textbf{Barbara Plank\textsuperscript{\faMountain\kern1pt\faRobot}} \quad
\textbf{Fosca Giannotti\textsuperscript{\faPizzaSlice}}
\\[8pt]
\textsuperscript{\faPizzaSlice}Scuola Normale Superiore, Italy \quad
\textsuperscript{\faGopuram}University of Pisa, Italy
\\
\textsuperscript{\faMountain}MaiNLP, LMU Munich, Germany \quad
\textsuperscript{\faRobot}Munich Center for Machine Learning, Germany \\
{\tt
\footnotesize
{\{\href{mailto:benedetta.muscato@sns.it}{\textcolor{black}{benedetta.muscato}}, \href{mailto:gizem.gezici@sns.it}{\textcolor{black}{ gizem.gezici}},\href{mailto:fosca.giannotti@sns.it}{\textcolor{black}{fosca.giannotti}}\}@sns.it,
\{\href{mailto:beiduo.chen@lmu.de}{\textcolor{black}{beiduo.chen}}, \href{mailto:b.plank@lmu.de}{\textcolor{black}{b.plank}}\}@lmu.de}}
}

\begin{document}
\maketitle

\begin{abstract}

Human disagreement is ubiquitous and well-known in labeling. However, variation in explanations, captured through token-level human rationales, remains far less explored.
At the same time, it is unclear how to best evaluate human labels and rationales---or even how to best aggregate rationales beyond majority vote---in light of this variation. Yet, rationales may provide additional insights into the richness of human reasoning, that may differ in style, values and interpretations---especially in subjective NLP tasks like hate speech detection.
In this work, we unify diverse models, training strategies, loss functions, and existing evaluation metrics under a single protocol by systematically re-implementing them across different label and rationale representation spaces. Classification metrics are organized around two key properties—predictive and distributional—while explainability metrics through three complementary dimensions: plausibility, faithfulness, and complexity.
In this unified supervision framework, we evaluate model behavior across classification and explainability metrics, as well as metric sensitivity to the choice of label (\textsc{hard} and \textsc{soft}) and rationale representation space (\textsc{hard}, \textsc{intermediate} and \textsc{soft}).
Results show that both hard and soft metrics favor softer representations, highlighting their effectiveness in capturing variation and the need to rethink evaluation in subjective NLP.

\end{abstract}

\noindent \textbf{Warning:} \textit{{This paper contains examples that may be offensive or upsetting.}}

\section{Introduction}

\begin{figure}[!ht]
    \centering
    \includegraphics[width=0.98\columnwidth]{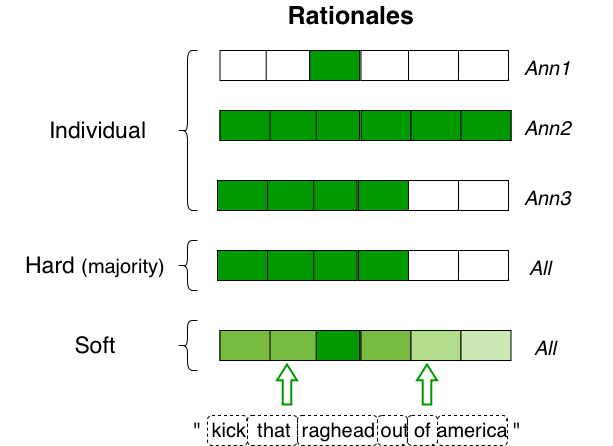} 

    \caption{
    \textbf{Variation in human rationale at the token level.} Annotators may highlight different tokens for the same prediction. \textsc{Hard} rationales keep only majority-selected tokens, collapsing disagreement, while \textsc{soft} rationales preserve graded token importance across annotators (color intensity).
    }
    \label{fig:intro_HEV}
\end{figure}

Designing Natural Language Processing (NLP) systems for subjective tasks, such as offensive or toxic language detection, has long been an active area of research \cite{founta2018large,sachdeva2022measuring}. However, traditional approaches often overlook the inherently subjective nature of these tasks, where individual backgrounds, beliefs, and values shape judgments about offensive content \cite{aroyo2015truth,sandri2023don}. These limitations are further reinforced by commonly used evaluation metrics, such as accuracy and F1-score, which focus on the majority label and fail to capture the nuances of ambiguous instances \cite{basile2021we}.

A recent line of work explicitly accounts for human disagreement in classification evaluation by modeling variation in the label space, better aligning with human judgments \cite{uma2020case, rizzi2024soft}. In particular, \citet{kurniawan2025training} propose an evaluation framework for the Human Label Variation (HLV) setting \cite{plank2022problem}, comparing predicted label distributions with empirical annotator distributions rather than fixed class labels. 
~\citet{hong2025litex} and ~\citet{jiang-etal-2023-ecologically} complementarily show that annotators may agree on labels but diverge in justifications, or vice versa, suggesting that explanations vary meaningfully across individuals (Figure~\ref{fig:intro_HEV}). Yet evaluating explanation quality remains challenging due to the lack of standardized protocols and agreed-upon metrics~\cite{dembinsky2025unifying,atanasova2020diagnostic,chen-etal-2025-rose}.

Calls for \textit{actionability} emphasize that the key question is not only \emph{how} well models can be explained, but also \emph{what} users can meaningfully do with those explanations \cite{orgad2026actionable}. This concern is particularly relevant in subjective and socially sensitive tasks, where clear explanations play a critical role in fostering trust and ensuring safety \cite{muscato2025perspectives}.

\paragraph{Contributions} Motivated by these challenges and the limited attention to explanation variation in prior work, which has mainly focused on label variation \cite{cabitza2023toward}, we rethink evaluation practices for classification and explainability in hate speech detection. Our contributions are as follows (Table~\ref{tab:comparison}):

\begin{itemize}
\setlength{\itemsep}{1pt}

\item We extend prior attention-supervision methods to support richer label and rationale spaces (\S\ref{sec:methodology}), introducing intermediate and soft supervision objectives.

\item We unify existing classification and explainability metrics under a single evaluation protocol (\S\ref{sec:evaluation}) for subjective tasks.

\item We evaluate model behavior across metrics, and assess metric sensitivity to different label and rationale representation spaces (\S\ref{sec:results_final}) using two attention-supervision methods, two transformer models, and two multilingual hate-speech datasets in English and Portuguese.

\end{itemize}

Our analysis shows that label and rationale representations influence evaluation, with softer representations generally improving classification and explainability performance.
This highlights the \textbf{challenge of modeling representations that reflect diverse human judgments} rather than a single truth, suggesting the need to \textbf{rethink evaluation in subjective NLP}.

\input{table/related_work}

\section{Related Work}
\label{sec:related_work} 

Explainability methods for NLP models include local explanations (e.g., feature attributions, attention maps, rationales) and global analyses based on probing or neuron-level inspection \cite{zhao2024explainability}. 
Rationale-based approaches produce either extractive token level rationales or abstractive free-text explanations \cite{mendez2024outputs}. 
We focus on self-explaining models using human token-level rationales to align model attention with human reasoning.

\paragraph{Explainable Hate Speech Detection}
A line of work leverages human rationales to improve explainability in hate speech detection. 
\textbf{HateXplain} \cite{mathew2021hatexplain}, \textbf{Masked Rationale Prediction} (\textbf{MRP}, \citealt{kim2022hate}), and \textbf{Supervised Rational Attention} (\textbf{SRA}, \citealt{eilertsen2025aligning}) incorporate human rationales into BERT-based models using the \textsc{HateXplain} dataset \cite{mathew2021hatexplain}, the first hate speech benchmark with token level rationale annotations. 

\citet{mathew2021hatexplain} aggregate annotator rationales into soft temperature-scaled attention distributions, while \citet{kim2022hate} propose \textbf{MRP}, a two-stage approach with masked-rationale pretraining followed by label fine-tuning. 
\citet{eilertsen2025aligning} introduce \textbf{SRA}, aligning attention with human rationales via a joint classification–alignment loss. 
Evaluation commonly follows the \textsc{ERASER} framework \cite{deyoung2020eraser}, combining classification metrics (\textsc{accuracy}, \textsc{macro-F1}) with explanation metrics for plausibility (\textsc{IoU-F1}, \textsc{Token-F1}, \textsc{AUPRC}), faithfulness (\textsc{comprehensiveness}, \textsc{sufficiency}), and complexity (\textsc{complexity}, \textsc{sparseness})~\cite{dhaini2025evalxnlp}.

\paragraph{Limitations and Open Challenges}
Attention supervision methods improve classification and explanation quality on benchmark datasets \cite{hartmann2022survey,strout2019human}, but assume a single ground-truth label and a 
rationale obtained via majority voting.
This approach collapses disagreement at both the label and explanation levels, particularly in subjective tasks such as hate speech detection, where annotators may reasonably disagree on labels and supporting tokens.
We argue that addressing such tasks requires classifiers that can \emph{model annotator disagreement} and generate variation-aware explanations reflecting \emph{diverse human judgments} rather than enforcing a single correct explanation.

\section{Methodology}
\label{sec:methodology}
This section presents an overview of our approach. We first introduce the label and rationale space (\S\ref{sec:representation}), followed by the models and training objectives (\S\ref{sec:cls_avg}).

\subsection{Label and Rationale Space}
\label{sec:representation}

We introduce representation settings for labels and rationales, where the latter depends on the label type (\textsc{hard} or \textsc{soft}); see Table~\ref{tab:intermediate_repr}.

\paragraph{\textsc{Hard} Label / \textsc{Hard} Rationale}
\label{par:hard_hard}
A single class obtained via majority voting, with \textbf{hard rationales}: binary token masks indicating whether a token is selected as evidence ($1$) or not ($0$) using a $0.5$ threshold. This standard setup (e.g., \textsc{HateXplain, MRP}, and \textsc{SRA}) serves as our main baseline.

\paragraph{\textsc{Hard} Label / \textsc{Intermediate} Rationale}
\label{par:hard_intermediate}

To capture annotator variability while keeping \textsc{hard} labels, we use \textbf{intermediate rationales}: \textbf{Union} (tokens selected by any annotator), \textbf{Random} (tokens randomly sampled), and \textbf{Full} (all tokens).

\paragraph{\textsc{Soft} Label / \textsc{Soft} Rationale}
\label{par:soft_soft}
Soft labels encode annotator label distributions, capturing disagreement across classes (e.g., $[0.67, 0.33]$). \textbf{Soft rationales} assign each token a continuous importance value to preserve subjectivity.

\input{table/intermediate_repr}

\subsection{Models and Training Objectives}
\label{sec:language_models_baseline}

We re-implement two rationale-based approaches, \textbf{\textsc{MRP}} \cite{kim2022hate} and \textbf{\textsc{SRA}} \cite{eilertsen2025aligning}, under different label and rationale representations and adopt the same hyperparameters.  Following prior work, we use a fixed random seed to ensure reproducibility. We focus on attention-based rationales as a practical and widely adopted proxy for language-model self-explanations.

Table~\ref{tab:comparison} summarizes the new setups; details are provided in Appendix~\ref{sec:appendix_loss}.

\paragraph{Loss}
For MRP, we follow the original paper and apply \textsc{CE} to \textsc{hard} labels and rationales, and \textsc{soft-CE}, \textsc{MSE}, and \textsc{KL-divergence} to their \textsc{soft} counterparts.
For SRA, the \textsc{hard} setting combines \textsc{CE} for classification with \textsc{MSE} for rationale supervision, as in the original work. In the \textsc{soft} setting, we experiment with several rationale losses: \textsc{soft-CE}~\cite{uma2020case}, \textsc{KL-divergence}, \textsc{inverse KL-divergence}~\cite{fornaciari2021beyond}, and \textsc{MSE}. These are matched with the same classification losses, except \textsc{MSE}, which is combined with \textsc{CE}. 

\vspace{-0.5em}
\paragraph{Attention Extraction}
\label{sec:cls_avg}

We extract token level attention from layer $l$, head $h$ as model-generated rationales. Let $A^{(l,h)} \in \mathbb{R}^{n \times n}$ denote the attention matrix for a sequence of length $n$. Following \citet{eilertsen2025aligning}, the \textbf{CLS} variant uses attention from the \texttt{[CLS]} query, $\hat{r}_i = A^{(l,h)}_{0,i}$ for $i=1,\dots,n$, while our \textbf{Avg.} variant averages attention across queries, $\hat{r}_i = \frac{1}{n}\sum_{j=1}^{n} A^{(l,h)}_{j,i}$. The resulting $\hat{\mathbf{r}} \in \mathbb{R}^n$ serves as the model rationale and is aligned with human rationales via the rationale loss.

\section{Evaluation Framework}
\label{sec:evaluation}
Our framework organizes existing metrics into two classification properties, \textit{predictive} and \textit{distributional}, and three explainability properties, \textit{plausibility}, \textit{faithfulness}, and \textit{complexity} (Table~\ref{tab:metrics_overview}).
\subsection{Classification Metrics}
\label{sec:classification_metrics}
We evaluate models with classification metrics capturing (i) predictive performance and (ii) alignment with the label distribution, depending on whether models predict a single label or a full distribution. See Appendix~\ref{sec:classification_metrics_appendix} for details.

\paragraph{Hard}
Standard classification metrics assume a single label per item. We report \textsc{Accuracy} and \textsc{macro-F1}, the class-averaged F1 score, balancing performance across classes~\cite{basile2021we,rodrigues2018deep}.

\begin{tcolorbox}[colback=blue!8, colframe=blue!40,
fontupper=\footnotesize, boxrule=0.2pt, arc=1pt, left=4pt, right=2pt, top=1pt, bottom=1pt, boxsep=0pt]
\pro\ Simple, interpretable, widely used.\\
\con\ Uses majority labels; ignores annotator disagreement.
\end{tcolorbox}

\paragraph{Soft}
Annotator-derived distributions $\mathbf{q}_i$ are compared with model predictions $\mathbf{p}_i$. We report \textsc{soft accuracy}, \textsc{soft macro-F1}, and \textsc{Jensen--Shannon Divergence} (\textsc{JSD}), capturing both diversity and consensus~\cite{kurniawan2025training}.

\textsc{Soft accuracy} correlates strongly with \textsc{JSD} and captures annotator disagreement, while \textsc{soft macro-F1} offers class-balanced scoring under label imbalance. Together, they extend hard metrics to probabilistic labels.
We adopt \textsc{JSD} as a symmetric and bounded alternative to KL divergence, following prior multi-perspective evaluation practices \cite{rizzi2024soft, kurniawan2025training,chen-etal-2024-seeing}, with advantages in stability and comparability across models and configurations.

\begin{tcolorbox}[colback=green!8, colframe=green!40,
fontupper=\footnotesize, boxrule=0.2pt, arc=1pt, left=2pt, right=2pt, top=1pt, bottom=1pt, boxsep=0pt]
\pro\ Capture annotator distributions, agreement, and class balance.\\
\con\ Require multi-annotator labels; unstable with sparse annotations.
\end{tcolorbox}

\subsection{Explainability Metrics}
\label{sec:explainability_metrics}

\input{table/taxonomy}
We evaluate rationales using \textsc{hard} and \textsc{soft} metrics capturing three key properties: \textbf{plausibility}, \textbf{faithfulness} as in~\cite{deyoung2020eraser}, and \textbf{complexity}~\cite{dhaini2025evalxnlp}. See Appendix~\ref{sec:explainability_metrics_appendix} for details.

\subsubsection{Plausibility} This property measures the alignment between model-generated and human rationales.

\paragraph{Hard} The standard rationale evaluation metrics are as follows. \textsc{IoU-F1} measures token overlap (intersection-over-union) using an IoU threshold (e.g., 50\%),
and \textsc{Token-F1} computes token level F1 between predicted and gold
rationale tokens~\cite{deyoung2020eraser}.

\begin{tcolorbox}[colback=blue!5, colframe=blue!40, fontupper=\small, boxrule=0.4pt, arc=2pt, left=4pt, right=4pt, top=2pt, bottom=2pt]
\pro\ Easy to compute; widely used in prior work.\\
\con\ Ignores graded importance, relies on fixed thresholds, and is token-based, missing semantic equivalence.
\end{tcolorbox}

\paragraph{Soft} This category evaluates continuous token-importance scores. \textsc{AUPRC} measures how well rationale tokens are ranked above non-rationale tokens via the area under the precision–recall curve, across thresholds~\cite{deyoung2020eraser}.

\begin{tcolorbox}[colback=green!5, colframe=green!40, fontupper=\small, boxrule=0.4pt, arc=2pt, left=4pt, right=4pt, top=2pt, bottom=2pt]
\pro\ Threshold-free; uses continuous scores.\\
\con\ Sensitive to class imbalance; dominated by positives.
\end{tcolorbox}

\subsubsection{Faithfulness}
Complementing plausibility, which measures alignment with human rationales, this property assesses whether rationales reflect the model’s decision process.

\paragraph{Hard} 
These metrics assess changes in the target class probability.
\textsc{Comprehensiveness} measures confidence drop after removing rationale tokens; larger drops suggest greater relevance of those tokens to the model's decision.
\textsc{Sufficiency} captures confidence drop when only rationale tokens are retained; smaller drops indicate that the rationale alone is sufficient to support the prediction~\cite{deyoung2020eraser}.

\begin{tcolorbox}[colback=blue!5, colframe=blue!40, fontupper=\small, boxrule=0.4pt, arc=2pt, left=4pt, right=4pt, top=1pt, bottom=1pt]
\pro\ Model-intrinsic; no human rationales required.\\
\con\ Require token binarization; incompatible with soft rationales.
\end{tcolorbox}

\paragraph{Soft} As an extension of hard metrics, the soft variant operates on continuous attention distributions, weighting tokens instead of applying binary masks and enabling finer-grained evaluation through probabilistic perturbations.~\cite{zhao2023incorporating}.

\textsc{Soft comprehensiveness} measures the confidence drop when tokens are perturbed according to attention weights; larger drops indicate higher relevance.
\textsc{Soft sufficiency} tests whether top-weighted tokens alone preserve the prediction; smaller changes indicate they capture most of the evidence.
\begin{tcolorbox}[colback=green!5, colframe=green!40,
fontupper=\small, boxrule=0.4pt, arc=1pt, left=2pt, right=4pt, top=1pt, bottom=1pt]
\pro\ No token binarization; supports soft rationales.\\
\con\ Sensitive to perturbations; less standardized.
\end{tcolorbox}

\subsubsection{Complexity}
This property measures rationales conciseness by encouraging importance to be concentrated on fewer tokens~\cite{dhaini2025evalxnlp}. We use two complementary, representation-agnostic metrics.
\textsc{Complexity}~\cite{bhatt2021evaluating} quantifies attribution spread via Shannon entropy; lower values indicate more focused explanations.
\textsc{Sparseness}~\cite{chalasani2020concise} measures attribution concentration using the Gini index\footnote{Distribution inequality (0 = uniform importance; 1 = all importance on a single token).}; higher values indicate more selective explanations.
\begin{tcolorbox}[colback=gray!8, colframe=gray!50,
fontupper=\small, boxrule=0.4pt, arc=2pt, left=4pt, right=4pt, top=2pt, bottom=2pt]
\pro\ Representation-agnostic.\\
\con\ Ignore semantic coherence; purely statistical.
\end{tcolorbox}

\section{Experimental Setup}
We evaluate our approach on hate speech datasets in English and Portuguese (\S\ref{sec:datasets}) with labels and token-level rationales. Our experiments use two BERT-based models (\S\ref{sec:models}) that leverage both signals during training. While recent work studies explanations in large language models (LLMs), we focus on encoder-based architectures for comparability with prior rationale-supervision methods. Code is available at \url{https://anonymous.4open.science/r/Variation_Human_Rationales-5F65}.

\label{sec:experimental_setup}
\subsection{Datasets}
\label{sec:datasets}

We use two datasets with human-annotated rationales: \textsc{HateXplain} \cite{mathew2021hatexplain} (English) and \textsc{HateBRXplain} \cite{salles2025hatebrxplain} (Portuguese), which exhibit different levels of annotator disagreement.
\textsc{HateXplain} contains 20K Twitter and Gab posts annotated by three annotators across three classes (\emph{normal}, \emph{offensive}, \emph{hate speech}), with inter-annotator agreement of $\alpha=0.42$ for labels and $\alpha=0.49$ for rationales\footnote{Krippendorff’s $\alpha$, measuring inter-annotator agreement over the corresponding annotations.}. \textsc{HateBRXplain} contains 7K Instagram comments about Brazilian politicians annotated by two experts for binary classification (\emph{offensive} vs.\ \emph{not offensive}), with agreement of $\alpha=0.75$ for labels and $\alpha=0.62$ for rationales. Both datasets use 80/10/10 train/validation/test splits with fixed random seeds. We select them due to the limited availability of benchmarks with token-level explanations for subjective tasks; alternative datasets often rely on free-text explanations or unusually large annotator pools per instance, limiting their comparability (\S\ref{sec:limitations}).

\input{table/hard_soft_agnostic_results_new}

\subsection{Models}
\label{sec:models}
We use \textbf{BERT-base-uncased}\footnote{\url{https://huggingface.co/google-bert/bert-base-uncased}} for \textsc{HateXplain} \cite{mathew2021hatexplain} and \textbf{BERTimbau-base}\footnote{\url{https://huggingface.co/neuralmind/bert-base-portuguese-cased}} for \textsc{HateBRXplain} \cite{salles2025hatebrxplain,vargas2022hatebr}, following \textsc{MRP} \cite{kim2022hate} and \textsc{SRA} \cite{eilertsen2025aligning}.
In these approaches, rationales supervise training by aligning model attention with human-provided token-level annotations via a regularization term $\alpha$.

On \textsc{HateXplain}, we evaluate \textbf{\textsc{HateXplain}} \cite{mathew2021hatexplain}, \textbf{\textsc{MRP}}, and \textbf{\textsc{SRA}}; on \textsc{HateBRXplain}, we evaluate \textbf{\textsc{MRP}} and \textbf{\textsc{SRA}}. We re-implement MRP and SRA with \textsc{hard} labels and \textsc{intermediate} rationales, and with \textsc{soft} labels and \textsc{soft} rationales, while also including the original \textsc{HateXplain} baseline. 
The \textbf{\textsc{HateXplain}} baseline uses \textsc{hard} labels with \textsc{soft} rationales, whereas \textbf{\textsc{MRP}} and \textbf{\textsc{SRA}} use \textsc{hard} labels with \textsc{hard} rationales (Table~\ref{tab:comparison}).
Additional details on the baseline configurations are given in Appendix~\ref{sec:appendix_loss}.

\subsection{Evaluation Metrics}

As discussed in~\S\ref{sec:evaluation}, we evaluate models on two dimensions: (i) classification performance and (ii) explanation quality.  
For \textsc{hard} labels, we report \colorbox{blue!8}{\textsc{Accuracy}} and \colorbox{blue!8}{\textsc{macro-F1}}; for \textsc{soft} labels, we use \colorbox{green!8}{\textsc{soft-Accuracy}}, \colorbox{green!8}{\textsc{soft-F1}}, and \colorbox{green!8}{\textsc{JSD}}. Explanations are assessed on three properties: \textbf{plausibility} (\colorbox{blue!8}{\textsc{IoU-F1}}, \colorbox{blue!8}{\textsc{Token F1}} for \textsc{hard}; \colorbox{green!8}{\textsc{AUPRC}} for \textsc{soft}), \textbf{faithfulness} (\colorbox{blue!8}{\textsc{Comprehensiveness}}, \colorbox{blue!8}{\textsc{Sufficiency}} for \textsc{hard}; continuous variants for \textsc{soft}), and \textbf{complexity} (\colorbox{gray!8}{\textsc{Complexity}}, \colorbox{gray!8}{\textsc{Sparsity}}, representation-agnostic).

\section{Results} 
\label{sec:results_final}

Table~\ref{tab:all_metrics} reports results for our best MRP and SRA configurations, which vary across label space, rationale space, and loss.
We evaluate both \textsc{[CLS]} and \textsc{Avg.} extraction methods (\S\ref{sec:cls_avg}): \textsc{Avg.} performs best on \textsc{HateXplain}, while \textsc{[CLS]} performs best on \textsc{HateBRXplain}. We then evaluate the best MRP and SRA configurations on both datasets across the \textsc{hard}, \textsc{intermediate}, and \textsc{soft} variants, reporting results under the corresponding evaluation metrics.

Our evaluation is organized around three dimensions: model behavior across classification and explainability metrics (\S\ref{sec:model_behaviour}), metric sensitivity to the choice of label/rationale space (\S\ref{sec:sensitivity}), and correlations among the metrics (\S\ref{sec:correlation}).

\subsection{Model Behavior}
\label{sec:model_behaviour}
To assess model behavior, we compare different model families (e.g., MRP and SRA) on \textsc{hard} and \textsc{soft} classification and explainability metrics.

For \textit{classification}, \textbf{MRP} achieves the strongest \textsc{hard} classification performance on both datasets, whereas \textbf{SRA} performs better on the \textsc{soft} classification metrics.
For \textit{explainability}, \textbf{SRA} achieves better results on both \textsc{hard} and \textsc{soft} \textbf{plausibility} (\textsc{IOU, Tok., \textsc{AUPRC}}) metrics on both datasets.
For \textbf{faithfulness} (\textsc{Comp.} and \textsc{Suff.}), \textbf{SRA} outperforms \textbf{MRP} on the \textsc{hard} metrics for both datasets, except for \textsc{Comp.} on \textsc{HateXplain}. In contrast, for \textsc{soft} \textbf{faithfulness}, \textbf{MRP} outperforms \textbf{SRA} on both datasets.
For \textbf{complexity} (\textsc{Cpx.} and \textsc{Spr.}), \textsc{SRA} is the best-performing model on the \textsc{agnostic} metrics.

\subsection{Metric Sensitivity} 
\label{sec:sensitivity} 

To assess metric sensitivity, we compare different label and rationale spaces using \textsc{hard} and \textsc{soft} classification and explainability metrics within each model family.
\textit{Overall, both \textsc{hard} and \textsc{soft} metrics favor softer (\textsc{soft} and \textsc{intermediate}) representations, which generally achieve the strongest performance on classification and explainability tasks. Plausibility metrics consistently benefit from softer representations, while for faithfulness, this trend holds primarily for \textsc{soft} faithfulness metrics.}

For \textit{classification}, using \textsc{hard} metrics, the \textsc{hard} and \textsc{soft} representation spaces are competitive for both \textbf{MRP} and \textbf{SRA} on \textsc{HateXplain}, whereas the \textsc{soft} representation
space performs best for both models on \textsc{HateBRXplain} 
For \textsc{soft} classification metrics, \textbf{MRP} is effective in both the \textsc{hard} and \textsc{soft} spaces, while \textbf{SRA} favors the \textsc{soft} space on both datasets.

Regarding \textit{explainability}, for \textbf{plausibility}, on \textsc{HateXplain}, \textbf{MRP} benefits more from the \textsc{soft} and \textsc{intermediate} spaces under \textsc{hard} metrics, while \textbf{SRA} remains relatively stable across representation spaces. Under \textsc{soft} metrics, \textbf{MRP} performs competitively in the \textsc{hard} and \textsc{soft} spaces, whereas \textbf{SRA} shows stronger performance in the \textsc{soft} and \textsc{intermediate} spaces. On \textsc{HateBRXplain}, plausibility scores are generally comparable across spaces for both models, although the \textsc{hard} space tends to yield the strongest \textsc{hard} plausibility performance for \textbf{SRA}.

For \textbf{faithfulness}, the \textsc{hard} representation space tends to perform strongly under \textsc{hard} faithfulness metrics, whereas the \textsc{soft} and \textsc{intermediate} spaces more often achieve similar or superior results under \textsc{soft} faithfulness metrics.
For \textbf{complexity} (\textsc{agnostic}), all representation spaces perform comparably on \textbf{SRA} models. In contrast, for \textbf{MRP}, the \textsc{soft} representation space achieves the best results on \textsc{HateExplain}, whereas the \textsc{hard}  outperforms the others on \textsc{HateBRXplain}.

\input{table/qualitative_analysis}

\subsection{Metric Correlation}
\label{sec:correlation}

After analyzing label and rationale representations across spaces (\textsc{hard--soft}), we examine correlations both across and within spaces (\textsc{hard--hard}, \textsc{soft--soft}) using statistical testing and Pearson's correlation.
We use statistical testing to assess whether observed differences are reliable, and Pearson correlation analysis to examine whether metrics vary together, thereby capturing similar or complementary patterns.

\paragraph{Statistical Analysis}
We assess significance with a paired $t$-test or Wilcoxon signed-rank test based on the Shapiro--Wilk normality test \citep{shapiro1965analysis}, apply FDR correction, and report Cohen’s $d$ and confidence intervals to account for the small sample size and effect size \citep{cohen2013statistical, serdar2021sample}. Further details are given in Appendix~\ref{sec:statistical_analysis}.

\paragraph{Correlation Analysis}

Across representation spaces (\textsc{hard--soft}), results on \textsc{HateXplain} reveal no significant differences between \textsc{hard} and \textsc{soft} spaces, with correlations ranging from negative to moderately positive across metrics. In contrast, for \textsc{HateBRXplain}, most metrics differ significantly across spaces. Correlations remain generally positive across metric pairs, particularly for \textsc{Tok}–\textsc{AUPRC}, whereas \textsc{Comp} shows almost no association and \textsc{Suff} exhibits a weak negative relationship.
These findings suggest that the two spaces are distinct yet partially aligned, likely because higher annotator agreement reduces noise. While genuine disagreement is valuable, noise may stem from annotation errors, unreliable annotators, or limitations of the annotation framework.

Within the same space (\textsc{hard--hard}, \textsc{soft--soft}), metric pairs exhibit consistent patterns across datasets. In the \textsc{hard} space, \textsc{Acc}–\textsc{F1} and \textsc{IoU}–\textsc{Tok} show strong positive correlations, whereas in the \textsc{soft} space, \textsc{Acc}–\textsc{F1} remains statistically significantly correlated in both datasets. Across both spaces, \textsc{Comp.} and \textsc{Suff.} are strongly negatively correlated, indicating a trade-off between \textsc{Comprehensivenss} and \textsc{Sufficiency}. Similarly, the complexity measures are negatively correlated in both datasets, significantly in \textsc{HateXplain} and perfectly, though not significantly, in \textsc{HateBRXplain}.

\section{Disagreeing Rationales}
Beyond quantitative evaluation, directly comparing explanations produced under different supervision signals can clarify whether the observed sensitivity comes from genuinely different explanations or from the metrics themselves.

To quantify rationale disagreement, we compute token-level \textsc{IoU} across annotators ($0.35$ and $0.67$ for \textsc{HateBRXplain} and \textsc{HateBRXplain} respectively) and between model variants. \textsc{SRA-hard} and \textsc{SRA-soft-loss} achieve $0.42$ (EN) and $0.51$ (PT), indicating moderate but non-trivial divergence. We also conduct a human evaluation study on \textsc{HateXplain}, revealing high subjectivity and low annotator agreement. Representative disagreed examples are presented in Appendix~\ref{sec:appendix_human_judgements}.

For a qualitative assessment of disagreed rationales, Table \ref{tab:qualitative_analysis} compares the \textsc{[H]SRA} baseline against our proposed soft variant \textsc{[S]SRA} on \textsc{HateXplain} (see Table \ref{tab:all_metrics}). The \textsc{soft} model produces more diverse rationales, highlighting a wider range of hateful terms (e.g., “illegal”, “refugees”, “idiot”), whereas the \textsc{hard} model focuses on a narrower subset. This suggests that \textsc{soft} supervision better captures rationale diversity.

\section{Discussion}
\label{sec:discussion}

\paragraph{No single model family dominates all metrics.}
MRP performs best on \textsc{hard} classification and \textsc{soft} faithfulness, whereas SRA achieves superior results on \textsc{soft} classification, \textsc{hard} plausibility, and \textsc{agnostic} complexity. Consequently, the choice of model should be guided by the specific evaluation objective.

\paragraph{Softer representations more effectively capture variation in labels and rationales.} Both \textsc{hard} and \textsc{soft} metrics favor \textsc{soft} and \textsc{intermediate} representations, indicating their effectiveness in capturing variation in labels and rationales.

\paragraph{Metrics within and across spaces might capture distinct patterns.}
Correlation analysis shows that metrics within and across spaces may capture different patterns rather than being directly interchangeable. Their utility depends on the evaluation goal: even metrics within the same space can compete as they assess different dimensions of the same property (e.g., faithfulness).
Aligning metric choice with the task and validating it with domain experts can help ensure more robust evaluation when modeling representations that reflect diverse human judgments in subjective NLP settings.

\section{Conclusion}
We standardized diverse models, training strategies, loss functions, and evaluation metrics from prior work in a unified supervision framework for assessing classification and explainability in light of variation-aware labels and rationales. 
In particular, we evaluated model behavior, metric sensitivity to the choice of label/rationale representation space and metric correlations.
Overall, our findings highlight the challenge of modeling representations that reflect diverse human judgments suggesting the need to rethink evaluation in subjective NLP.

\newpage
\section*{Limitations}
\label{sec:limitations}

While our analysis offers insights into the challenges of classification and explainability evaluation under variability in human rationales, several limitations persist.

First, our study is constrained by the limited availability of benchmark datasets for subjective NLP tasks that provide multiple labels and rationales per instance. Existing resources such as SBIC \cite{sap2020social} and Gaze4Hate \cite{alacam2024eyes} differ in key respects: SBIC offers free-text explanations, whereas our work focuses on token-level rationales optimized via a joint classification–alignment objective. Gaze4Hate, by contrast, incorporates a substantially larger annotator pool and additional experimental complexity. Consequently, we center our experiments on HateXplain and HateBRXplain, which are widely adopted and enable more direct comparison with prior work.

Second, the scarcity of suitable datasets limits the generalizability of our findings to other subjective NLP tasks characterized by variation in human rationales. In particular, the absence of resources that jointly provide multiple labels and token-level rationales hinders a broader assessment of the applicability of our conclusions.

Third, our experiments rely on BERT-based models to ensure comparability with established rationale-supervised baselines such as MRP and SRA. While this choice facilitates controlled comparisons, it constrains the scope of our analysis. Future work could examine how variation in human rationales interacts with explanations produced by large autoregressive or instruction-tuned language models.

Fourth, although we benchmark attention-based explanations (CLS/Avg.), attention weights may not faithfully capture model reasoning~\citep{jain2019attention}. Future work will extend the framework to alternative attribution methods, such as gradient-based (e.g. Integrated Gradients, Layer Integrated Gradients) and perturbation-based approaches (e.g. LIME and SHAP), enabling controlled comparisons across explanation families and testing whether divergences persist beyond attention-based explanations.

Finally, our focus on token-level rationales constrains the range of explanation formats considered. Although rationales offer a structured and convenient supervision signal for training and evaluation, they constitute only one form of explanation. In many real-world settings, free-text explanations may provide richer and more interpretable justifications for model predictions. Recent datasets include such natural language explanations, and integrating them with large language models capable of generating explanations represents a promising direction for future research on modeling and evaluating variation in human explanations.

\section*{Ethics Statement}

Hate speech detection is an inherently complex task, as annotators' judgments and the rationales they provide are shaped by factors such as cultural background, personal experience, and contextual interpretation \cite{davani2024d3code}. In this work, we treat variation in both labels and rationales not as noise but as a meaningful reflection of diverse human perspectives. Rather than collapsing annotator disagreement into a single ground truth, our approach deliberately leverages this variation by averaging across annotators to produce soft labels and soft rationales. We consider this particularly important given that marginalized communities are disproportionately affected both by hateful content and by the errors of automated moderation systems \cite{davidson2019racial}, and that imposing a single majority viewpoint risks silencing the very perspectives most needed to identify subtle forms of harm. As language models become increasingly ubiquitous in content moderation and everyday applications, ensuring that their decisions are transparent, human-aligned, and inclusive of diverse viewpoints represents both a technical objective and an ethical commitment. 

\paragraph{Use of AI Assistants}
The authors acknowledge the use of ChatGPT solely for correcting grammatical errors, enhancing the coherence of the final manuscript.



\appendix

\section{Appendix}
\label{sec:appendix}

\subsection{Baselines and Loss Functions}
\label{sec:appendix_loss}

We detail below the hyperparameters of our trained models and their loss functions used for both label prediction and rationale supervision across the MRP and SRA baselines.

\paragraph{Hyperameters} 

We adopt the same hyperparameters as the baselines for both MRP and SRA. Table~\ref{tab:hyperparams} and \ref{tab:hyperparams_MRP} summarize all training details.

\input{table/hyperparams}

\subsubsection{Label Loss}
\label{par:label_loss_appendix}

\paragraph{Cross-Entropy (CE).}
Used for hard label training in both MRP and SRA. Given a model output logit vector $\mathbf{z} \in \mathbb{R}^C$ and the ground-truth class $y$:

\begin{equation}
    \mathcal{L}_{\text{CE}} = -\log \frac{\exp(z_y)}{\sum_{c=1}^{C} \exp(z_c)}
\end{equation}

\paragraph{Soft Cross-Entropy (Soft-CE).}
Used for soft label training in MRP. Let $\mathbf{p} = \text{softmax}(\mathbf{z})$ be the predicted distribution and $\mathbf{q}$ the soft target distribution (HJD):
\begin{equation}
    \mathcal{L}_{\text{soft-CE}} = -\sum_{c=1}^{C} q_c \log p_c
\end{equation}

\paragraph{KL-Divergence Regular (KL-Reg).}
Used for soft label training in SRA. Measures the divergence from the target distribution $\mathbf{q}$ to the predicted distribution $\mathbf{p}$:
\begin{equation}
    \mathcal{L}_{\text{KL}} = \sum_{c=1}^{C} q_c \log_2 \frac{q_c}{p_c}
\end{equation}

\paragraph{Inverse KL-Divergence (KL-Inv).}
Reverses the direction of the divergence, where the predicted distribution drives the loss:
\begin{equation}
    \mathcal{L}_{\text{KL-inv}} = \sum_{c=1}^{C} p_c \log_2 \frac{p_c}{q_c}
\end{equation}

\subsubsection{Rationale Loss}
\label{par:exp_loss_appendix}

In SRA, the rationale loss aligns the model's attention vector $\hat{\mathbf{r}} \in \mathbb{R}^n$ with human rationale annotations $\mathbf{r} \in \mathbb{R}^n$, where $n$ is the sequence length. All losses below are computed only over non-padded positions, masked by the attention mask $\mathbf{m} \in \{0,1\}^n$.

\paragraph{Mean Squared Error (MSE).}
Used in hard-label SRA settings. For soft rationales, both $\hat{\mathbf{r}}$ and $\mathbf{r}$ are normalized to sum to one before comparison:
\begin{equation}
    \mathcal{L}_{\text{MSE}} = \frac{1}{|\mathcal{V}|} \sum_{i \in \mathcal{V}} \left( \hat{r}_i - r_i \right)^2
\end{equation}
where $\mathcal{V} = \{i : m_i = 1\}$ is the set of valid (non-padded) token positions.

\paragraph{KL-Divergence Regular (KL-Reg).}
Applied after normalizing both attention and rationale vectors to valid probability distributions over non-padded tokens:
\begin{equation}
    \mathcal{L}_{\text{KL-rat}} = \sum_{i \in \mathcal{V}} \tilde{r}_i \log_2 \frac{\tilde{r}_i}{\tilde{a}_i}
\end{equation}

\paragraph{Inverse KL-Divergence (KL-inverse).}
The reverse direction, where the model's normalized attention drives the divergence:
\begin{equation}
    \mathcal{L}_{\text{KL-inv-rat}} = \sum_{i \in \mathcal{V}} \tilde{a}_i \log_2 \frac{\tilde{a}_i}{\tilde{r}_i}
\end{equation}

\subsubsection{Total Training Objective}

For MRP, the total loss is simply the label loss ($\mathcal{L}_{\text{CE}}$ or $\mathcal{L}_{\text{soft-CE}}$). For SRA, the total loss combines label prediction and rationale supervision:
\begin{equation}
    \mathcal{L}_{\text{total}} = \mathcal{L}_{\text{label}} + \alpha \cdot \mathcal{L}_{\text{rationale}}
\end{equation}
where $\alpha$ is a weighting hyperparameter controlling the influence of rationale supervision on the training objective.

\subsection{Evaluation Metrics}
\label{sec:appendix_metrics}

As discussed in the previous sections, we evaluate our models based on classification and explainability performance.

\subsubsection{Classification Metrics}
\label{sec:classification_metrics_appendix}

\paragraph{Accuracy.}
The fraction of correctly predicted instances:
\begin{equation}
    \text{Accuracy} = \frac{1}{N} \sum_{i=1}^{N} \mathbb{1}[\hat{y}_i = y_i]
\end{equation}
\paragraph{Macro-F1.}
The unweighted mean of per-class F1 scores, where for each class $c$:
\begin{equation}
\begin{aligned}
    \text{Precision}_c &= \frac{\text{TP}_c}{\text{TP}_c + \text{FP}_c}, \\
    \text{Recall}_c &= \frac{\text{TP}_c}{\text{TP}_c + \text{FN}_c}
\end{aligned}
\end{equation}
\begin{equation}
\begin{aligned}
    \text{Macro-F1} &= \frac{1}{C} \sum_{c=1}^{C} \text{F1}_c
\end{aligned}
\end{equation}

\paragraph{Soft-Accuracy.}
For soft labels, accuracy is computed as the probability mass assigned by the model to the ground-truth distribution. Let $\hat{\mathbf{p}}_i$ be the predicted distribution and $\mathbf{q}_i$ the soft target for instance $i$:
\begin{equation}
    \text{Soft-Acc} = \frac{1}{N} \sum_{i=1}^{N} \sum_{c=1}^{C} q_{i,c} \cdot \hat{p}_{i,c}
\end{equation}

\paragraph{Soft-F1.}
Extends macro-F1 to soft labels by weighting per-class contributions by the target distribution rather than hard assignments.

\paragraph{Jensen--Shannon Divergence (JSD).}
Measures the symmetric divergence between the predicted and target soft-label distributions:
\begin{equation}
    \mathbf{m}_i = \frac{1}{2}(\hat{\mathbf{p}}_i + \mathbf{q}_i)
\end{equation}
\begin{equation}
    \text{JSD}(\hat{\mathbf{p}}_i \| \mathbf{q}_i) = \frac{1}{2} D_{\text{KL}}(\hat{\mathbf{p}}_i \| \mathbf{m}_i) + \frac{1}{2} D_{\text{KL}}(\mathbf{q}_i \| \mathbf{m}_i)
\end{equation}

\subsubsection{Explainability Metrics}
\label{sec:explainability_metrics_appendix}

\paragraph{Plausibility.}
Plausibility measures how well model-generated rationales align with human annotations.

\textit{IoU-F1 (hard).} Intersection-over-Union between the predicted rationale token set $\hat{R}$ and the human rationale set $R$:
\begin{equation}
    \text{IoU} = \frac{|\hat{R} \cap R|}{|\hat{R} \cup R|}, \quad
    \text{IoU-F1} = \frac{2 \cdot \text{IoU}}{1 + \text{IoU}}
\end{equation}

\textit{Token-F1 (hard).} Standard token-level F1 treating each rationale token as a binary classification:
\begin{equation}
    \text{Token-P} = \frac{|\hat{R} \cap R|}{|\hat{R}|}, \quad
    \text{Token-R} = \frac{|\hat{R} \cap R|}{|R|}
\end{equation}
\begin{equation}
    \text{Token-F1} = \frac{2 \cdot \text{Token-P} \cdot \text{Token-R}}{\text{Token-P} + \text{Token-R}}
\end{equation}

\textit{AUPRC (soft).} Area Under the Precision--Recall Curve, treating the soft human rationale as a continuous relevance signal and the model attention weights as predicted scores. Higher AUPRC indicates better alignment between the model's attention distribution and the degree of annotator agreement at each token position.

\paragraph{Faithfulness.}
Faithfulness measures whether the rationale tokens are truly influential for the model's prediction.

\textit{Comprehensiveness (hard).} Measures the drop in predicted probability when rationale tokens are removed:
\begin{equation}
    \text{Comp} = p(\hat{y} \mid \mathbf{x}) - p(\hat{y} \mid \mathbf{x} \setminus R)
\end{equation}
where $p(\hat{y} \mid \mathbf{x})$ is the model's confidence in its prediction given the full input, and $p(\hat{y} \mid \mathbf{x} \setminus R)$ is the confidence after removing rationale tokens. Higher values indicate more faithful rationales.

\textit{Sufficiency (hard).} Measures how well the rationale tokens alone support the prediction:
\begin{equation}
    \text{Suff} = p(\hat{y} \mid \mathbf{x}) - p(\hat{y} \mid R)
\end{equation}
where $p(\hat{y} \mid R)$ is the model's confidence when only rationale tokens are retained. Lower values indicate that the rationale is sufficient for the prediction.

\textit{Fuzzy Comprehensiveness and Sufficiency (soft).} For soft rationales, token removal is replaced by soft masking: each token's embedding is scaled by $(1 - r_i)$ for comprehensiveness and by $r_i$ for sufficiency, where $r_i \in [0,1]$ is the soft rationale score. This preserves the continuous nature of the rationale signal.

\paragraph{Complexity.}
Complexity metrics evaluate how concise and focused the generated rationales are.

\textit{Agnostic Complexity.} The fraction of tokens selected as rationale out of all valid (non-padded) tokens:
\begin{equation}
    \text{Complexity} = \frac{|\hat{R}|}{|\mathcal{V}|}
\end{equation}
Lower values indicate more concise rationales.

\textit{Sparsity.} The complement of complexity, measuring the fraction of non-rationale tokens:
\begin{equation}
    \text{Sparsity} = 1 - \text{Complexity} = 1 - \frac{|\hat{R}|}{|\mathcal{V}|}
\end{equation}
Higher values indicate sparser, more focused explanations.

\subsection{Statistical Analysis}
\label{sec:statistical_analysis}
We report the overall results for the best configurations on both \textsc{HateXplain} and \textsc{HateBRXplain} datasets in Table \ref{tab:all_metrics}.
We assess statistical significance using the paired $t$-test \cite{field2013discovering} or the Wilcoxon signed-rank test \citep{wilcoxon1945individual}, a non-parametric paired test used when the normality assumption does not hold. When the pairwise differences do not satisfy the normality assumption, as confirmed by the Shapiro--Wilk test \citep{shapiro1965analysis}, we use the Wilcoxon signed-rank test instead of the paired $t$-test. We then apply False Discovery Rate (FDR) correction to adjust $p$-values for multiple comparisons.

In addition to statistical significance, we report Cohen's $d$ as an effect size measure and confidence intervals to provide both effect-size information and uncertainty about the estimates. While $p$-values indicate whether observed differences are unlikely to be due to chance, effect sizes capture the magnitude of those differences; together, they provide complementary evidence for interpreting the results. We follow Cohen's conventional thresholds of $0.2$, $0.5$, and $0.8$ for small, medium, and large effects, respectively \citep{cohen2013statistical}.
Because the sample size is small (\(n \approx 10\)), only relatively large effects are likely to reach statistical significance, so effect sizes are especially important \cite{serdar2021sample, sullivan2012using}.
Across all results, statistically significant differences are associated with large effect sizes.

\begin{table}[h]
\centering
\small
\setlength{\tabcolsep}{4.0pt}
\renewcommand{\arraystretch}{1.1}
\begin{tabular}{@{}llrrc@{}}
\toprule
\textbf{Pair} & \textbf{Test} & \textbf{Effect size} & \textbf{95\% CI} & \textbf{$p$-value} \\
\midrule
\multicolumn{5}{l}{\textit{Hard vs.\ Soft}} \\
\cmidrule(l){1-5}
ACC        & $t$-test   & 0.40 & {[-.006, .026]}    & 0.269 \\
F1         & Wilcoxon   & 0.44 & {[.002, .062]}     & 0.269 \\
IoU--AUPRC & $t$-test   & 0.79 & {[-.360, -.055]}   & 0.104 \\
Tok--AUPRC & $t$-test   & 0.37 & {[-.176, .031]}    & 0.269 \\
Comp       & $t$-test   & 0.61 & {[.006, .150]}     & 0.175 \\
Suff       & $t$-test   & 1.06 & {[-.295, -.091]}   & 0.051 \\
\bottomrule
\end{tabular}
\caption{\textsc{HateXplain}: \textsc{Hard} vs.\ \textsc{Soft} metric comparison. Hard ACC, F1, Comp, and Suff are compared with their soft counterparts.
$p$-values are FDR-corrected at $\alpha = 0.05$; the results are not statistically significant.}
\label{tab:stat_hard_vs_soft_hatex}
\end{table}

\begin{table}[!t]
\centering
\small
\setlength{\tabcolsep}{4pt}
\renewcommand{\arraystretch}{1.1}
\begin{tabular}{@{}llrrc@{}}
\toprule
\textbf{Pair} & \textbf{Test} & \textbf{Effect size} & \textbf{99.5\% CI} & \textbf{$p$-value}  \\
\midrule
\multicolumn{5}{l}{\textit{Hard vs.\ Hard}} \\
\cmidrule(l){1-5}
ACC-F1    & $t$-test  & 1.84 & {[}.006, .016{]}  &  0.001*\\
IoU-Tok   & Wilcoxon  & 0.89 & {[-}.207, -.038{]} & 0.003* \\
Comp-Suff & Wilcoxon  & 0.85 & {[}.235, .468{]}  & 0.004* \\
\midrule
\multicolumn{5}{l}{\textit{Soft vs.\ Soft}} \\
\cmidrule(l){1-5}
ACC-F1    & Wilcoxon  & 0.89 & {[.004, .059]}   & 0.004* \\
Comp-Suff & Wilcoxon  & 0.93 & {[-.099, .135]}   & 0.250\\
\midrule
\multicolumn{5}{l}{\textit{Agnostic vs.\ Agnostic}} \\
\cmidrule(l){1-5}
Cpx-Spr   & Wilcoxon  & 0.89 & {[-.924, -.597]}  & 0.002* \\
\bottomrule
\end{tabular}
\caption{Within-space metric comparisons on the \textsc{HateXplain} dataset. $p$-values are FDR-corrected at $\alpha = 0.005$; $^{*}$ indicates significance.}
\label{tab:stat_within_label}
\end{table}

\begin{table}[h]
\centering
\small
\setlength{\tabcolsep}{4.0pt}
\renewcommand{\arraystretch}{1.1}
\begin{tabular}{@{}llrrc@{}}
\toprule
\textbf{Pair} & \textbf{Test} & \textbf{Effect size} & \textbf{95\% CI} & \textbf{$p$-value} \\
\midrule
\multicolumn{5}{l}{\textit{Hard vs.\ Soft}} \\
\cmidrule(l){1-5}
ACC        & Wilcoxon  & 0.53 & {[-.005, .062]}    & 0.155 \\
F1         & Wilcoxon  & 0.89 & {[.028, .193]}     & 0.011* \\
IoU--AUPRC & $t$-test  & 1.65 & {[-.363, -.162]}   & 0.007* \\
Tok--AUPRC & $t$-test  & 1.26 & {[-.183, -.059]}   & 0.011* \\
Comp       & $t$-test  & 1.16 & {[-.316, -.101]}   & 0.012* \\
Suff       & $t$-test  & 0.22 & {[-.063, .138]}    & 0.531 \\
\bottomrule
\end{tabular}
\caption{\textsc{HateBRXplain}: \textsc{Hard} vs.\ \textsc{Soft} metric comparison. Hard ACC, F1, Comp, and Suff are compared against their soft counterparts. $p$-values are FDR-corrected at $\alpha = 0.05$; $^{*}$ indicates significance.}
\label{tab:stat_hard_vs_soft_hatebr}
\end{table}

\begin{table}[h]
\centering
\small
\setlength{\tabcolsep}{4.0pt}
\renewcommand{\arraystretch}{1.1}
\begin{tabular}{@{}llrrc@{}}
\toprule
\textbf{Pair} & \textbf{Test} & \textbf{Effect size} & \textbf{95\% CI} & \textbf{$p$-value}  \\
\midrule
\multicolumn{5}{l}{\textit{Hard vs.\ Hard}} \\
\cmidrule(l){1-5}
ACC-F1    & Wilcoxon  & 0.32 & {[-}.002, .004{]}  & 1.000 \\
IoU-Tok   & $t$-test  & 1.90 & {[-}.187, -.095{]} & 0.001* \\
Comp-Suff & $t$-test  & 0.56 & {[-}.020, .214{]}  & 0.198 \\
\midrule
\multicolumn{5}{l}{\textit{Soft vs.\ Soft}} \\
\cmidrule(l){1-5}
ACC-F1    & Wilcoxon  & 0.89 & {[.022, .158]}   & 0.016* \\
Comp-Suff & Wilcoxon  & 0.85 & {[.251, .487]}   & 0.016*\\
\midrule
\multicolumn{5}{l}{\textit{Agnostic vs.\ Agnostic}} \\
\cmidrule(l){1-5}
Cpx-Spr   & Wilcoxon  & 0.49 & {[-.853, -.108]}  & 0.156 \\
\bottomrule
\end{tabular}
\caption{Within-space metric comparisons on the \textsc{HateBRXplain} dataset. $p$-values are FDR-corrected at $\alpha = 0.05$; $^{*}$ indicates significance.}
\label{tab:stat_within_label_hatebr}
\end{table}

\subsection{Correlation Analysis}

We report in the following section the Pearson's correlation \cite{benesty2009pearson} results across \textsc{hard--hard}, \textsc{soft--soft}, \textsc{hard--soft} and \textsc{agnostic--agnostic} representations for both \textsc{HateXplain} (Table \ref{tab:pearson_hatexplain}) and \textsc{HateBRXplain} (Table \ref{tab:pearson_hatebrxplain}).
\begin{table}[!t]
\centering
\small
\setlength{\tabcolsep}{4pt}
\renewcommand{\arraystretch}{1.1}
\begin{tabular}{@{}lr@{}}
\toprule
\textbf{Pair} & \textbf{Pearson $r$} \\
\midrule
\multicolumn{2}{l}{\textit{Hard vs.\ Hard — Classification}} \\
\cmidrule(l){1-2}
Acc–F1             & 0.92 \\
\midrule
\multicolumn{2}{l}{\textit{Hard vs.\ Hard — Explainability}} \\
\cmidrule(l){1-2}
IoU F1–Token F1    & 0.78 \\
Compr–Suff         & $-$0.52 \\
\midrule
\multicolumn{2}{l}{\textit{Soft vs.\ Soft — Classification}} \\
\cmidrule(l){1-2}
Soft Acc–Soft F1   & 0.66   \\

\midrule
\multicolumn{2}{l}{\textit{Soft vs.\ Soft — Explainability}} \\
\cmidrule(l){1-2}
AUPRC–Soft Compr   & $-$0.28 \\
AUPRC–Soft Suff    & 0.57    \\
Soft Compr–Soft Suff & $-$0.67\\
\midrule
\multicolumn{2}{l}{\textit{Hard vs.\ Soft — Classification}} \\
\cmidrule(l){1-2}
Acc–Soft Acc       & $-$0.28 \\
F1–Soft F1         & $-$0.58 \\
\midrule
\multicolumn{2}{l}{\textit{Hard vs.\ Soft — Explainability}} \\
\cmidrule(l){1-2}
IoU F1–AUPRC       & 0.01    \\
Token F1–AUPRC     & 0.50    \\
Compr–Soft Compr   & 0.41    \\
Suff–Soft Suff     & $-$0.46 \\
\midrule
\multicolumn{2}{l}{\textit{Agnostic vs.\ Agnostic}} \\
\cmidrule(l){1-2}
Cpx–Spr            & $-$0.78 \\
\bottomrule
\end{tabular}
\caption{HateXplain Pearson's correlation coefficients across metric spaces.}
\label{tab:pearson_hatexplain}
\end{table}

\begin{table}[!t]
\centering
\small
\setlength{\tabcolsep}{4pt}
\renewcommand{\arraystretch}{1.1}
\begin{tabular}{@{}lr@{}}
\toprule
\textbf{Pair} & \textbf{Pearson $r$} \\
\midrule
\multicolumn{2}{l}{\textit{Hard vs.\ Hard — Classification}} \\
\cmidrule(l){1-2}
Acc–F1             & 1.00 \\
\midrule
\multicolumn{2}{l}{\textit{Hard vs.\ Hard — Explainability}} \\
\cmidrule(l){1-2}
IoU F1–Token F1    & 0.90 \\
Compr–Suff         & $-$0.74 \\
\midrule
\multicolumn{2}{l}{\textit{Soft vs.\ Soft — Classification}} \\
\cmidrule(l){1-2}
Soft Acc–Soft F1   & 0.93 \\
\midrule
\multicolumn{2}{l}{\textit{Soft vs.\ Soft — Explainability}} \\
\cmidrule(l){1-2}
AUPRC–Soft Compr   & $-$0.12 \\
AUPRC–Soft Suff    & $-$0.11 \\
Soft Compr–Soft Suff & $-$0.90\\
\midrule
\multicolumn{2}{l}{\textit{Hard vs.\ Soft — Classification}} \\
\cmidrule(l){1-2}
Acc–Soft Acc       & 0.51    \\
F1–Soft F1         & 0.31    \\
\midrule
\multicolumn{2}{l}{\textit{Hard vs.\ Soft — Explainability}} \\
\cmidrule(l){1-2}
IoU F1–AUPRC       & 0.35    \\
Token F1–AUPRC     & 0.65    \\
Compr–Soft Compr   & 0.01    \\
Suff–Soft Suff     & $-$0.21 \\
\midrule
\multicolumn{2}{l}{\textit{Agnostic vs.\ Agnostic}} \\
\cmidrule(l){1-2}
Cpx–Spr            & $-$1.00\\
\bottomrule
\end{tabular}
\caption{\textsc{HateBRXplain} Pearson's correlation coefficients across metric spaces on the second dataset.}
\label{tab:pearson_hatebrxplain}
\end{table}

\subsection{Human Judgments}
\label{sec:appendix_human_judgements}

To examine the alignment between automatic metrics and human evaluation, we conduct a study on forty instances from the English HateXplain dataset with five annotators of diverse socio-demographic backgrounds (age, gender, nationality). Instances are selected by shuffling across four conditions: correct label + correct rationale, wrong label + wrong rationale, correct label + wrong rationale, and wrong label + correct rationale. We also show two illustrative examples where annotators mostly disagreed on the various judged explainability dimensions.

Annotators are shown the instance, the prediction from the best soft model (\textsc{SRA-soft-loss}), and its rationale. We clarified the intent of our study and informed annotators that their judgments are collected solely for research purposes.

The instruction provided to the annotators are reported:

\paragraph{Annotator Instructions}

You will evaluate the model-generated explanations (highlighted rationale spans) along three dimensions: \textbf{plausibility}, \textbf{faithfulness}, and \textbf{complexity}.

\medskip
\noindent For each instance, you will see:
\begin{itemize}[nosep]
    \item the original text with highlighted tokens shown in \textbf{bold} inside \textlangle\,\textrangle
    \item the model's predicted label
\end{itemize}

\noindent Score each dimension from 1 (worst) to 5 (best).

\medskip

\paragraph{Evaluation Dimensions}

\begin{enumerate}
    \item \textbf{Plausibility (1--5)}

    Do the highlighted spans make sense from a human perspective? Would you select similar words or phrases to justify the given label?
    \begin{itemize}[nosep]
        \item 1 = Highlights are completely irrelevant
        \item 5 = Highlights closely match what a human would select
    \end{itemize}

    \item \textbf{Faithfulness (1--5)}

    Do the highlighted spans accurately reflect what likely influenced the model's prediction? If the highlighted tokens were removed, would the prediction likely change?
    \begin{itemize}[nosep]
        \item 1 = Highlights do not align with the prediction at all
        \item 5 = Highlights fully capture the model's decision basis
    \end{itemize}

    \item \textbf{Complexity (1--5)}

    Does the explanation make the model's decision easier to understand? Is it clear, concise, and helpful for identifying potential errors or biases?
    \begin{itemize}[nosep]
        \item 1 = Provides no useful insight
        \item 5 = Clearly communicates the model's reasoning
    \end{itemize}
\end{enumerate}

\paragraph{Qualitative analysis of judged instances}

We complement our human evaluation study with a small qualitative analysis aimed at examining whether specific instances exhibit systematically higher or lower inter-annotator agreement on the evaluation of explainability quality. We select two representative examples from the participant questionnaire: one with \textbf{high agreement} and one with \textbf{high disagreement} across explainability dimensions. For each case, we report the model rationales alongside the distribution of human ratings for plausibility, faithfulness, and complexity.

\begin{table}[ht]
\centering
\small
\begin{tabular}{p{0.95\columnwidth}}
\toprule
\textbf{High-agreement instance} \\
\midrule
\textbf{Text:} \textit{``It is true that if I am a moderate muzzy and I do not stand against the extremists within my religion I am a worthless dog who deserves the title of muzzy terrorist period.''} \\[4pt]
\textbf{Rationales:} true, if, muzzy, do, extremists, my, religion, dog, who, deserves, terrorist. \\[4pt]
\textbf{Judgments:} \\
Plausibility: 80\% score 4, 20\% score 3. \\
Faithfulness: 80\% score 5, 20\% score 3. \\
Complexity: 100\% score 4. \\
\bottomrule
\end{tabular}
\caption{Instance with high inter-annotator agreement across all explainability dimensions.}
\label{tab:high_agreement}
\end{table}

This example shows consistent judgments across evaluators, indicating strong agreement on the perceived alignment with the input (plausibility), the relevance of the rationale (faithfulness), and the perceived simplicity of the rationale (complexity).

\begin{table}[ht]
\centering
\small
\begin{tabular}{p{0.95\columnwidth}}
\toprule
\textbf{Low-agreement instance} \\
\midrule
\textbf{Text:} \textit{``Girls are joining Boy Scouts now better be so they can bring them sandwiches like real women.''} \\[4pt]
\textbf{Rationales:} joining, boyscouts, better, be, bring, real, women. \\[4pt]
\textbf{Judgments:} \\
Plausibility: 40\% score 4, 20\% score 3, 20\% score 2, 20\% score 1. \\
Faithfulness: 60\% score 4, 20\% score 3, 20\% score 2. \\
Complexity: 40\% score 4, 40\% score 2, 20\% score 1. \\
\bottomrule
\end{tabular}
\caption{Instance with high inter-annotator disagreement across explainability dimensions.}
\label{tab:low_agreement}
\end{table}

In contrast, this instance exhibits higher variability across annotators, reflecting disagreement in how the rationales are interpreted and evaluated. Notably, although the selected tokens capture salient lexical cues from the input, annotators differ in whether these cues constitute meaningful explanations. This results in inconsistent judgments across plausibility, faithfulness, and complexity, suggesting that perceived alignment, explanation correctness, and perceived simplicity of the rationales relative to the input are highly subjective.
This behavior is consistent with the inherently ambiguous nature of the instance, which relies on implicit and context-dependent hateful language that is open to interpretation and therefore prone to annotator disagreement.

\end{document}

%% file: table/related_work.tex
\begin{table*}[!t]
\centering
\tiny
\renewcommand{\arraystretch}{0.2}
\setlength{\tabcolsep}{1.5pt}
\resizebox{0.9\textwidth}{!}{%
\begin{tabular}{@{} l c c c c @{}}
\toprule
 & HateXplain & MRP & SRA & \textbf{Ours (Unified Supervision Framework)} \\
\midrule
\textit{Label} &
\textsc{Hard} &
\textsc{Hard} &
\textsc{Hard} &
\textsc{Hard}, \textsc{\textbf{Soft}} \\
\midrule
\textit{Rationale} &
\textsc{Soft} &
\textsc{Hard} &
\textsc{Hard} &
\textsc{\textsc{\textbf{Intermediate}}, Soft} \\
\midrule
\textit{Training} &
\makecell{$\textsuperscript{$\star$}$Joint training} &
\makecell{$\textsuperscript{$\star\star$}$Mask then classify} &
\makecell{$\textsuperscript{$\star$}$Joint training} &
\makecell{$\textsuperscript{$\star$}$Joint training \& $\textsuperscript{$\star\star$}$Mask then classify} \\
\midrule
\textit{Label Loss} &
CE&
CE &
CE &
\makecell{CE, \textbf{Soft CE}, \textbf{KL}, \textbf{MSE}} \\
\midrule
\textit{Rationale Loss} &
CE &
CE &
MSE &
\makecell{CE, \textbf{Soft CE}, \textbf{KL}, MSE} \\

\bottomrule
\end{tabular}%
}
\caption{Overview of prior explainable hate speech methods and our unified supervision framework, with our new setups shown in \textbf{bold}. Prior work varies across \textbf{label/rationale spaces}, \textbf{training strategies} (\textsuperscript{$\star$}joint training with attention loss; \textsuperscript{$\star\star$}masked-rationale pretraining followed by label fine-tuning), and \textbf{loss functions} (CE = Cross-Entropy, MSE = Mean Squared Error, KL = Kullback--Leibler divergence). Our framework \textbf{standardizes} these dimensions and \textbf{extends} supervision across label and rationale spaces, supporting \textsc{hard} and \textsc{soft} labels, \textsc{intermediate} and \textsc{soft} rationales, and multiple supervision losses under a single evaluation protocol.}
\label{tab:comparison}
\end{table*}

%% file: table/intermediate_repr.tex
\begin{table}[!h]
\centering
\small
\resizebox{0.90\linewidth}{!}{%
\begin{tabular}{@{}l@{}}
\toprule
\textbf{Rationale Representations} \\
\midrule
\textbf{Ann.~1:} what a \colorbox{yellow}{pakis} favorite clothing brand? \colorbox{yellow}{jihadidas} \\[2pt]
\textbf{Ann.~2:} what a \colorbox{yellow}{pakis} favorite clothing brand? \colorbox{yellow}{jihadidas} \\[2pt]
\textbf{Ann.~3:} what a pakis favorite clothing brand? \colorbox{yellow}{jihadidas} \\
\midrule
\midrule
\textbf{\textsc{Hard:}} what a \colorbox{yellow}{pakis} favorite clothing brand? \colorbox{yellow}{jihadidas} \\[4pt]
\midrule
\textbf{\textsc{Union}}: what a \colorbox{yellow}{pakis} favorite clothing brand? \colorbox{yellow}{jihadidas} \\[2pt]
\textbf{\textsc{Random}}: what a \colorbox{yellow}{pakis} \colorbox{yellow}{favorite} clothing brand? \colorbox{yellow}{jihadidas} \\[2pt]
\textbf{\textsc{Full}}: \colorbox{yellow}{what a pakis favorite clothing brand? jihadidas} \\[4pt]
\midrule
\textbf{\textsc{Soft}}: what a \colorbox{yellow!33}{pakis} favorite clothing brand? \colorbox{yellow}{jihadidas} \\
\bottomrule
\end{tabular}%
}
\caption{Example of rationale representations as \textsc{hard}, \textsc{intermediate} (\textsc{union, random, full}), and \textsc{soft} for the example: \textit{What a pakis favorite clothing brand? Jihadidas.}}
\label{tab:intermediate_repr}
\end{table}

%% file: table/taxonomy.tex
\begin{table*}[!t]
\centering
\resizebox{0.80\linewidth}{!}{
\setlength{\tabcolsep}{3.0pt}
\renewcommand{\arraystretch}{0.7}
\begin{tabular}{@{}llll@{}}
\toprule
\textbf{Label Space} & \textbf{Property} & \textbf{Metric} & \textbf{Reference} \\
\midrule
\multicolumn{4}{@{}l}{\textbf{\textit{\enspace Classification}}} \\
\midrule
\cellcolor{blue!8} & \cellcolor{blue!8}
  & \cellcolor{blue!8} \textsc{Accuracy}\, ($\up$) & \cellcolor{blue!8} \\
\cellcolor{blue!8} \multirow{-2}{*}{\textsc{Hard}} & \cellcolor{blue!8} \multirow{-2}{*}{Predictive Performance}
  & \cellcolor{blue!8} \textsc{Macro-F1} ($\up$) & \cellcolor{blue!8} \multirow{-2}{*}{\citet{sokolova2009systematic}} \\
\cmidrule(lr){1-4}
\cellcolor{green!8} & \cellcolor{green!8}
  & \cellcolor{green!8} \textsc{Soft Accuracy} ($\up$) & \cellcolor{green!8} \\
\cellcolor{green!8} & \cellcolor{green!8} \multirow{-2}{*}{Predictive Performance}
  & \cellcolor{green!8} \textsc{Soft Macro-F1} ($\up$) & \cellcolor{green!8} \multirow{-2}{*}{\citet{kurniawan2025training}} \\[4pt]
\cellcolor{green!8} \multirow{-3}{*}{\textsc{Soft}} & \cellcolor{green!8} Distribution Alignment
  & \cellcolor{green!8} JSD ($\down$) & \cellcolor{green!8} \citet{lin2002divergence} \\
\midrule
\multicolumn{4}{@{}l}{\textbf{\textit{\enspace Explainability}}} \\
\midrule
\cellcolor{blue!8} & \cellcolor{blue!8}
  & \cellcolor{blue!8} \textsc{IoU-F1} ($\up$) & \cellcolor{blue!8} \\
\cellcolor{blue!8} & \cellcolor{blue!8} \multirow{-2}{*}{Plausibility}
  & \cellcolor{blue!8} \textsc{Token-F1} ($\up$) & \cellcolor{blue!8} \multirow{-2}{*}{\citet{deyoung2020eraser}} \\[4pt]
\cellcolor{blue!8} & \cellcolor{blue!8}
  & \cellcolor{blue!8} \textsc{Comprehensiveness}\ ($\up$) & \cellcolor{blue!8} \\
\cellcolor{blue!8} \multirow{-4}{*}{\textsc{Hard}} & \cellcolor{blue!8} \multirow{-2}{*}{Faithfulness}
  & \cellcolor{blue!8} \textsc{Sufficiency} ($\down$) & \cellcolor{blue!8} \multirow{-2}{*}{\citet{deyoung2020eraser}} \\[3pt]
\cmidrule(lr){1-4}
\cellcolor{green!8} & \cellcolor{green!8} Plausibility
  & \cellcolor{green!8} \textsc{AUPRC} ($\up$) & \cellcolor{green!8} \citet{deyoung2020eraser} \\[3pt]
\cellcolor{green!8} & \cellcolor{green!8}
  & \cellcolor{green!8} \textsc{Soft Comprehensiveness}\ ($\up$) & \cellcolor{green!8} \\
\cellcolor{green!8} \multirow{-3}{*}{\textsc{Soft}} & \cellcolor{green!8} \multirow{-2}{*}{Faithfulness}
  & \cellcolor{green!8} \textsc{Soft Sufficiency}\ ($\down$) & \cellcolor{green!8} \multirow{-2}{*}{\citet{zhao2023incorporating}} \\[3pt]
\cmidrule(lr){1-4}
\cellcolor{gray!8} & \cellcolor{gray!8}
  & \cellcolor{gray!8} \textsc{Complexity} ($\down$) & \cellcolor{gray!8} \citet{bhatt2021evaluating} \\
\cellcolor{gray!8} \multirow{-2}{*}{\textsc{Agnostic}} & \cellcolor{gray!8} \multirow{-2}{*}{Complexity}
  & \cellcolor{gray!8} \textsc{Sparseness} ($\up$) & \cellcolor{gray!8} \citet{chalasani2020concise} \\
\bottomrule
\end{tabular}
}
\caption{Evaluation metrics by label space and property. 
\textbf{Classification}: \textsc{Hard} metrics use majority labels; \textsc{Soft} metrics use annotator label distributions. 
\textbf{Explainability}: \textsc{Hard} metrics require binarized rationales; \textsc{Soft} metrics operate on continuous attribution distributions; \textsc{Agnostic} metrics are representation-independent.}
\label{tab:metrics_overview}
\end{table*}

%% file: table/hard_soft_agnostic_results_new.tex
\begin{table*}[h!]
\centering
\small
\setlength{\tabcolsep}{0.3pt}
\newcommand{\cg}[1]{\cellcolor{green!8}{#1}}
\newcommand{\cy}[1]{\cellcolor{gray!8}{#1}}
\newcommand{\cb}[1]{\cellcolor{blue!8}{#1}}
\begin{tabular}{l|cc|cccc|ccc|ccc|cc}
\toprule
& \multicolumn{6}{c|}{\cellcolor{blue!8}\textbf{\textsc{Hard}}} & \multicolumn{6}{c|}{\cellcolor{green!8}\textbf{\textsc{Soft}}} & \multicolumn{2}{c}{\cellcolor{gray!8}\textbf{\textsc{Agnostic}}} \\
\cmidrule(lr){2-7} \cmidrule(lr){8-13} \cmidrule(lr){14-15}
& \multicolumn{2}{c|}{\cb{Classif.}} & \multicolumn{4}{c|}{\cb{Explain.}} & \multicolumn{3}{c|}{\cg{Classif.}} & \multicolumn{3}{c|}{\cg{Explain.}} & \multicolumn{2}{c}{\cy{Explain.}} \\
\textbf{[Rep.] \& Approach} &
\cb{\textsc{Acc.}$\uparrow$} &
\cb{\textsc{F1}$\uparrow$} &
\cb{\textsc{IoU}$\uparrow$} &
\cb{\textsc{Tok.}$\uparrow$} &
\cb{\textsc{Comp.}$\uparrow$} &
\cb{\textsc{Suff.}$\downarrow$} &
\cg{\textsc{Acc.}$\uparrow$} &
\cg{\textsc{F1}$\uparrow$} &
\cg{\textsc{JSD}$\downarrow$} &
\cg{\textsc{AUPRC}$\uparrow$} &
\cg{\textsc{Comp.}$\uparrow$} &
\cg{\textsc{Suff.}$\downarrow$} &
\cy{\textsc{Cpx.}$\downarrow$} &
\cy{\textsc{Spr.}$\uparrow$} \\
\midrule
\multicolumn{15}{c}{\textit{\textbf{HateXplain}}} \\
\midrule
\textsc{[{S}] HateXplain}
  & \cb{.698} & \cb{\underline{.687}} & \cb{.120} & \cb{.411} & \cb{.424} & \cb{.160}
  & \cg{\underline{.695}} & \cg{.686} & \cg{.161} & \cg{.626} & \cg{.421} & \cg{.411}
  & \cy{.018} & \cy{.911} \\
\midrule
\textsc{[H] MRP}
  & \cb{\textbf{.704}} & \cb{\textbf{.699}} & \cb{.141} & \cb{.504} & \cb{\textbf{.479}} & \cb{.067}
  & \cg{.688} & \cg{.611} & \cg{.125} & \cg{.745} & \cg{\textbf{.652}} & \cg{\underline{.145}}
  & \cy{.080} & \cy{.630} \\
\textsc{[IU] MRP}
  & \cb{\underline{.703}} & \cb{.681} & \cb{.495} & \cb{.534} & \cb{.421} & \cb{.099}
  & \cg{.668} & \cg{.596} & \cg{.134} & \cg{.239} & \cg{\underline{.578}} & \cg{.197}
  & \cy{.114} & \cy{.515} \\
\textsc{[S] MRP-mse}
  & \cb{.694} & \cb{.684} & \cb{.301} & \cb{.373} & \cb{.260} & \cb{.293}
  & \cg{.680} & \cg{.608} & \cg{.125} & \cg{.316} & \cg{.475} & \cg{\textbf{.092}}
  & \cy{.076} & \cy{.624} \\
\textsc{[S] MRP-klreg}
  & \cb{\textbf{.704}} & \cb{\underline{.687}} & \cb{.498} & \cb{.552} & \cb{.370} & \cb{.095}
  & \cg{.655} & \cg{.591} & \cg{.139} & \cg{.435} & \cg{.513} & \cg{.165}
  & \cy{.065} & \cy{.652} \\
\midrule
\textsc{[H] SRA}
  & \cb{.696} & \cb{.682} & \cb{\underline{.539}} & \cb{\underline{.651}} & \cb{.417} & \cb{\underline{-.013}}
  & \cg{.644} & \cg{.634} & \cg{.180} & \cg{.753} & \cg{.442} & \cg{.283}
  & \cy{\textbf{.009}} & \cy{\textbf{.971}} \\
\textsc{[IU]SRA}
  & \cb{.674} & \cb{.662} & \cb{.484} & \cb{.612} & \cb{\underline{.428}} & \cb{-.031}
  & \cg{.692} & \cg{.685} & \cg{.124} & \cg{\underline{.764}} & \cg{.509} & \cg{.289}
  & \cy{.013} & \cy{.942} \\
\textsc{[IR] SRA}
  & \cb{.659} & \cb{.659} & \cb{\textbf{.557}} & \cb{.635} & \cb{.380} & \cb{\textbf{-.010}}
  & \cg{.676} & \cg{.675} & \cg{.131} & \cg{.681} & \cg{.510} & \cg{.231}
  & \cy{\underline{.012}} & \cy{.948} \\
\textsc{[S] SRA-soft-loss}
  & \cb{.687} & \cb{.678} & \cb{.372} & \cb{.474} & \cb{.332} & \cb{\underline{.012}}
  & \cg{\textbf{.702}} & \cg{\textbf{.697}} & \cg{\textbf{.118}} & \cg{.725} & \cg{.417} & \cg{.337}
  & \cy{.058} & \cy{\underline{.968}} \\
\textsc{[S] SRA-klreg}
  & \cb{.684} & \cb{.673} & \cb{.530} & \cb{\textbf{.681}} & \cb{.406} & \cb{-.033}
  & \cg{.694} & \cg{\underline{.690}} & \cg{\underline{.120}} & \cg{\textbf{.798}} & \cg{.386} & \cg{.334}
  & \cy{.017} & \cy{.901} \\
\midrule
\multicolumn{15}{c}{\textit{\textbf{HateBRXplain}}} \\
\midrule
\textsc{[H] MRP}
  & \cb{.902} & \cb{.902} & \cb{\underline{.447}} & \cb{.552} & \cb{.158} & \cb{.046}
  & \cg{.849} & \cg{.616} & \cg{.056} & \cg{.607} & \cg{.123} & \cg{.284}
  & \cy{.192} & \cy{.157} \\
\textsc{[IU] MRP}
  & \cb{.899} & \cb{.899} & \cb{.356} & \cb{.479} & \cb{.218} & \cb{.204}
  & \cg{.832} & \cg{.619} & \cg{.068} & \cg{.502} & \cg{\textbf{.606}} & \cg{\textbf{.082}}
  & \cy{.219} & \cy{-.024} \\
\textsc{[S] MRP}
  & \cb{.714} & \cb{.697} & \cb{.465} & \cb{.521} & \cb{.299} & \cb{.115}
  & \cg{.820} & \cg{.648} & \cg{.066} & \cg{.578} & \cg{.459} & \cg{\underline{.085}}
  & \cy{.209} & \cy{.069} \\
\textsc{[S] MRP-mse}
  & \cb{\textbf{.918}} & \cb{\textbf{.918}} & \cb{.347} & \cb{.466} & \cb{.190} & \cb{.258}
  & \cg{.828} & \cg{.631} & \cg{.064} & \cg{.526} & \cg{.553} & \cg{.130}
  & \cy{.209} & \cy{.067} \\
\midrule
\textsc{[H] SRA}
  & \cb{.904} & \cb{.906} & \cb{\textbf{.716}} & \cb{\textbf{.745}} & \cb{\textbf{.454}} & \cb{\textbf{-.036}}
  & \cg{.877} & \cg{.876} & \cg{.061} & \cg{\textbf{.831}} & \cg{.430} & \cg{.108}
  & \cy{\underline{.004}} & \cy{.979} \\
\textsc{[IU]SRA}
  & \cb{.897} & \cb{.897} & \cb{.321} & \cb{.556} & \cb{.259} & \cb{.080}
  & \cg{.877} & \cg{.876} & \cg{.063} & \cg{.763} & \cg{.539} & \cg{.097}
  & \cy{\underline{.004}} & \cy{\underline{.980}} \\
\textsc{[IR] SRA}
  & \cb{.904} & \cb{.904} & \cb{.373} & \cb{.535} & \cb{.268} & \cb{.073}
  & \cg{\underline{.880}} & \cg{\underline{.880}} & \cg{.061} & \cg{.694} & \cg{.533} & \cg{.095}
  & \cy{\underline{.004}} & \cy{\underline{.980}} \\
\textsc{[S] SRA-soft-loss}
  & \cb{.896} & \cb{.896} & \cb{.105} & \cb{.322} & \cb{.142} & \cb{.289}
  & \cg{\textbf{.886}} & \cg{\textbf{.885}} & \cg{\underline{.049}} & \cg{.663} & \cg{\underline{.554}} & \cg{.110}
  & \cy{\textbf{.002}} & \cy{\textbf{.989}} \\
\textsc{[S] SRA-klreg}
  & \cb{\underline{.909}} & \cb{\underline{.909}} & \cb{.429} & \cb{\underline{.655}} & \cb{\underline{.388}} & \cb{\underline{.040}}
  & \cg{.876} & \cg{.876} & \cg{\textbf{.048}} & \cg{\underline{.814}} & \cg{.427} & \cg{.120}
  & \cy{.005} & \cy{.972} \\
\bottomrule
\end{tabular}
\caption{Top-performing \textsc{hard}, \textsc{intermediate}, and \textsc{soft} configurations on \textsc{HateXplain} and \textsc{HateBRXplain}, evaluated with \textsc{hard}, \textsc{soft}, and \textsc{agnostic} metrics. Metric groups are color-coded by representation: \colorbox{blue!8}{hard}, \colorbox{green!8}{soft}, and \colorbox{gray!8}{agnostic}. 
\textsc{HateXplain}, \textsc{MRP}, and \textsc{SRA} denote baseline families, where suffixes specify a shared loss function applied jointly to both labels and rationales. When no suffix is provided, we use the original loss combination defined for each baseline (\S\ref{sec:language_models_baseline}). [H] = hard label/rationale, [S] = soft label/rationale, [IU] = hard label + intermediate union rationale, and [IR] = hard label + intermediate random rationale. For \textsc{HateXplain}, [S] denotes hard labels with soft rationales. Best results are in \textbf{bold}, second-best are \underline{underlined}.}

\label{tab:all_metrics}
\end{table*}

%% file: table/qualitative_analysis.tex
\begin{table*}[!ht]
\centering
\resizebox{\textwidth}{!}{%
\begin{tabular}{@{}lp{12.0cm}r@{}}
\toprule
\multicolumn{3}{@{}l}{\textbf{Instance:} \textit{``those that are for illegal immigration and islamic culture refugees are idiot liberal progressives and globalists shills''}} \\
\midrule
\textbf{Annotator} & \textbf{Human Rationale} & \textbf{Annotator Label} \\
\midrule
\textbf{Ann.1} & illegal, immigration, islamic, liberal, progressives, globalists & HS \\[3pt]
\textbf{Ann.2} & immigration, islamic, culture, refugees, idiot, liberal, progressives, globalists & OF \\[3pt]
\textbf{Ann.3} & illegal, immigration, islamic, progressives, globalists & HS \\
\midrule
\textbf{Model} & \textbf{Model Rationale} & \textbf{Model Label} \\
\midrule
\textbf{[H] SRA} & illegal, immigration, islamic, liberal, progressives, globalists & HS \\[3pt]
\textbf{[S] SRA} & illegal, immigration, islamic, culture, refugees, idiot, liberal, progressives, globalists & OF \\
\bottomrule
\end{tabular}
}
\caption{Example from the HateXplain dataset showing rationales predicted by different models (SRA \textsc{hard} and \textsc{soft}) and human annotators. HS = hate speech, OF = offensive. \textsc{[H]} denotes hard label/rationale and \textsc{[S]} denotes soft label/rationale. For SRA \textsc{soft}, we report the most probable label; the predicted distribution is [0.0, 0.7, 0.3].}
\label{tab:qualitative_analysis}
\end{table*}

%% file: table/hyperparams.tex
\begin{table}[ht]
\centering
\small
\resizebox{\linewidth}{!}{
\begin{tabular}{lcc}
\toprule
\textbf{Datasets} & \textsc{HateXplain} & \textsc{HateBRXplain} \\
\midrule
Optimizer & AdamW & AdamW \\
Learning rate & $2\times10^{-5}$ & $1\times10^{-5}$ \\
Batch size & 16 & 8 \\
Max seq.\ length & 128 & 512 \\
Epochs & 5 & 5 \\
$\alpha$ (align.\ weight) & 10.0 & 10.0 \\
Attention\_layer & 8 & 8 \\
Attention\_head & 7 & 7 \\

\bottomrule
\end{tabular}}
\caption{Hyperparameters for \textbf{SRA} across datasets.}
\label{tab:hyperparams}
\end{table}

\begin{table}[ht]
\centering
\small
\resizebox{\linewidth}{!}{
\begin{tabular}{lcc}
\toprule
\textbf{Stages} & \textsc{1-Rationale} & \textsc{2-Classification} \\
\midrule
Optimizer & AdamW & AdamW \\
Learning rate & $5\times10^{-5}$ & $2\times10^{-5}$ \\
Batch size & 16 & 16 \\
Max seq.\ length & 512 & 512 \\
Epochs & 5 & 5 \\
Attention\_layer & 12 & 12 \\
Attention\_head & 12 & 12 \\
Num\_labels & 2 & 3 \\
Mask\_ratio & 0.5 & - \\
\bottomrule
\end{tabular}}
\caption{Hyperparameters for \textbf{MRP} for both datasets.}
\label{tab:hyperparams_MRP}
\end{table}

%% file: main.bbl
\begin{thebibliography}{48}
\providecommand{\natexlab}[1]{#1}

\bibitem[{Alacam et~al.(2024)Alacam, Hoeken, and Zarrie{\ss}}]{alacam2024eyes}
{\"O}zge Alacam, Sanne Hoeken, and Sina Zarrie{\ss}. 2024.
\newblock Eyes don’t lie: Subjective hate annotation and detection with gaze.
\newblock In \emph{Proceedings of the 2024 Conference on Empirical Methods in Natural Language Processing}, pages 187--205.

\bibitem[{Aroyo and Welty(2015)}]{aroyo2015truth}
Lora Aroyo and Chris Welty. 2015.
\newblock Truth is a lie: Crowd truth and the seven myths of human annotation.
\newblock \emph{AI magazine}, 36(1):15--24.

\bibitem[{Atanasova et~al.(2020)Atanasova, Simonsen, Lioma, and Augenstein}]{atanasova2020diagnostic}
Pepa Atanasova, Jakob~Grue Simonsen, Christina Lioma, and Isabelle Augenstein. 2020.
\newblock A diagnostic study of explainability techniques for text classification.
\newblock In \emph{Proceedings of the 2020 conference on empirical methods in natural language processing (EMNLP)}, pages 3256--3274.

\bibitem[{Basile et~al.(2021)Basile, Fell, Fornaciari, Hovy, Paun, Plank, Poesio, and Uma}]{basile2021we}
Valerio Basile, Michael Fell, Tommaso Fornaciari, Dirk Hovy, Silviu Paun, Barbara Plank, Massimo Poesio, and Alexandra Uma. 2021.
\newblock We need to consider disagreement in evaluation.
\newblock In \emph{Proceedings of the 1st workshop on benchmarking: past, present and future}, pages 15--21.

\bibitem[{Benesty et~al.(2009)Benesty, Chen, Huang, and Cohen}]{benesty2009pearson}
Jacob Benesty, Jingdong Chen, Yiteng Huang, and Israel Cohen. 2009.
\newblock Pearson correlation coefficient.
\newblock In \emph{Noise reduction in speech processing}, pages 1--4. Springer.

\bibitem[{Bhatt et~al.(2021)Bhatt, Weller, and Moura}]{bhatt2021evaluating}
Umang Bhatt, Adrian Weller, and Jos{\'e}~MF Moura. 2021.
\newblock Evaluating and aggregating feature-based model explanations.
\newblock In \emph{Proceedings of the Twenty-Ninth International Conference on International Joint Conferences on Artificial Intelligence}, pages 3016--3022.

\bibitem[{Cabitza et~al.(2023)Cabitza, Campagner, and Basile}]{cabitza2023toward}
Federico Cabitza, Andrea Campagner, and Valerio Basile. 2023.
\newblock Toward a perspectivist turn in ground truthing for predictive computing.
\newblock In \emph{Proceedings of the AAAI Conference on Artificial Intelligence}, volume~37, pages 6860--6868.

\bibitem[{Chalasani et~al.(2020)Chalasani, Chen, Chowdhury, Wu, and Jha}]{chalasani2020concise}
Prasad Chalasani, Jiefeng Chen, Amrita~Roy Chowdhury, Xi~Wu, and Somesh Jha. 2020.
\newblock Concise explanations of neural networks using adversarial training.
\newblock In \emph{International conference on machine learning}, pages 1383--1391. PMLR.

\bibitem[{Chen et~al.(2025)Chen, Peng, Korhonen, and Plank}]{chen-etal-2025-rose}
Beiduo Chen, Siyao Peng, Anna Korhonen, and Barbara Plank. 2025.
\newblock \href {https://doi.org/10.18653/v1/2025.findings-acl.562} {A rose by any other name: {LLM}-generated explanations are good proxies for human explanations to collect label distributions on {NLI}}.
\newblock In \emph{Findings of the Association for Computational Linguistics: ACL 2025}, pages 10777--10802, Vienna, Austria. Association for Computational Linguistics.

\bibitem[{Chen et~al.(2024)Chen, Wang, Peng, Litschko, Korhonen, and Plank}]{chen-etal-2024-seeing}
Beiduo Chen, Xinpeng Wang, Siyao Peng, Robert Litschko, Anna Korhonen, and Barbara Plank. 2024.
\newblock \href {https://doi.org/10.18653/v1/2024.findings-emnlp.842} {``seeing the big through the small'': Can {LLM}s approximate human judgment distributions on {NLI} from a few explanations?}
\newblock In \emph{Findings of the Association for Computational Linguistics: EMNLP 2024}, pages 14396--14419, Miami, Florida, USA. Association for Computational Linguistics.

\bibitem[{Cohen(2013)}]{cohen2013statistical}
Jacob Cohen. 2013.
\newblock \emph{Statistical power analysis for the behavioral sciences}.
\newblock routledge.

\bibitem[{Davani et~al.(2024)Davani, Diaz, Baker, and Prabhakaran}]{davani2024d3code}
Aida Davani, Mark Diaz, Dylan Baker, and Vinodkumar Prabhakaran. 2024.
\newblock D3code: Disentangling disagreements in data across cultures on offensiveness detection and evaluation.
\newblock In \emph{Proceedings of the 2024 Conference on Empirical Methods in Natural Language Processing}, pages 18511--18526.

\bibitem[{Davidson et~al.(2019)Davidson, Bhattacharya, and Weber}]{davidson2019racial}
Thomas Davidson, Debasmita Bhattacharya, and Ingmar Weber. 2019.
\newblock Racial bias in hate speech and abusive language detection datasets.
\newblock In \emph{Proceedings of the third workshop on abusive language online}, pages 25--35.

\bibitem[{Dembinsky et~al.(2025)Dembinsky, Lucieri, Frolov, Najjar, Watanabe, and Dengel}]{dembinsky2025unifying}
David Dembinsky, Adriano Lucieri, Stanislav Frolov, Hiba Najjar, Ko~Watanabe, and Andreas Dengel. 2025.
\newblock Unifying vxai: a systematic review and framework for the evaluation of explainable ai.
\newblock \emph{arXiv preprint arXiv:2506.15408}.

\bibitem[{DeYoung et~al.(2020)DeYoung, Jain, Rajani, Lehman, Xiong, Socher, and Wallace}]{deyoung2020eraser}
Jay DeYoung, Sarthak Jain, Nazneen~Fatema Rajani, Eric Lehman, Caiming Xiong, Richard Socher, and Byron~C Wallace. 2020.
\newblock Eraser: A benchmark to evaluate rationalized nlp models.
\newblock In \emph{Proceedings of the 58th annual meeting of the association for computational linguistics}, pages 4443--4458.

\bibitem[{Dhaini et~al.(2025)Dhaini, Hussain, Zaradoukas, and Kasneci}]{dhaini2025evalxnlp}
Mahdi Dhaini, Kafaite~Zahra Hussain, Efstratios Zaradoukas, and Gjergji Kasneci. 2025.
\newblock Evalxnlp: A framework for benchmarking post-hoc explainability methods on nlp models.
\newblock \emph{arXiv preprint arXiv:2505.01238}.

\bibitem[{Eilertsen et~al.(2025)Eilertsen, Bj{\o}rgfinsd{\'o}ttir, Vargas, and Ramezani-Kebrya}]{eilertsen2025aligning}
Brage Eilertsen, R{\o}skva Bj{\o}rgfinsd{\'o}ttir, Francielle Vargas, and Ali Ramezani-Kebrya. 2025.
\newblock Aligning attention with human rationales for self-explaining hate speech detection.
\newblock In \emph{Proceedings of the AAAI Conference on Artificial Intelligence (AAAI)}.
\newblock To appear.

\bibitem[{Field(2013)}]{field2013discovering}
Andy Field. 2013.
\newblock \emph{Discovering Statistics Using IBM SPSS Statistics}.
\newblock SAGE.

\bibitem[{Fornaciari et~al.(2021)Fornaciari, Uma, Paun, Plank, Hovy, and Poesio}]{fornaciari2021beyond}
Tommaso Fornaciari, Alexandra Uma, Silviu Paun, Barbara Plank, Dirk Hovy, and Massimo Poesio. 2021.
\newblock Beyond black \& white: Leveraging annotator disagreement via soft-label multi-task learning.
\newblock In \emph{Proceedings of the 2021 Conference of the North American Chapter of the Association for Computational Linguistics: Human Language Technologies}, pages 2591--2597.

\bibitem[{Founta et~al.(2018)Founta, Djouvas, Chatzakou, Leontiadis, Blackburn, Stringhini, Vakali, Sirivianos, and Kourtellis}]{founta2018large}
Antigoni Founta, Constantinos Djouvas, Despoina Chatzakou, Ilias Leontiadis, Jeremy Blackburn, Gianluca Stringhini, Athena Vakali, Michael Sirivianos, and Nicolas Kourtellis. 2018.
\newblock Large scale crowdsourcing and characterization of twitter abusive behavior.
\newblock In \emph{Proceedings of the international AAAI conference on web and social media}, volume~12.

\bibitem[{Hartmann and Sonntag(2022)}]{hartmann2022survey}
Mareike Hartmann and Daniel Sonntag. 2022.
\newblock A survey on improving nlp models with human explanations.
\newblock In \emph{Proceedings of the first workshop on learning with natural language supervision}, pages 40--47.

\bibitem[{Hong et~al.(2025)Hong, Chen, Peng, de~Marneffe, and Plank}]{hong2025litex}
Pingjun Hong, Beiduo Chen, Siyao Peng, Marie-Catherine de~Marneffe, and Barbara Plank. 2025.
\newblock Litex: A linguistic taxonomy of explanations for understanding within-label variation in natural language inference.
\newblock In \emph{Proceedings of the 2025 Conference on Empirical Methods in Natural Language Processing}, pages 34053--34073.

\bibitem[{Jain and Wallace(2019)}]{jain2019attention}
Sarthak Jain and Byron~C Wallace. 2019.
\newblock Attention is not explanation.
\newblock In \emph{Proceedings of the 2019 Conference of the North American Chapter of the Association for Computational Linguistics: Human Language Technologies, Volume 1 (Long and Short Papers)}, pages 3543--3556.

\bibitem[{Jiang et~al.(2023)Jiang, Tan, and de~Marneffe}]{jiang-etal-2023-ecologically}
Nan-Jiang Jiang, Chenhao Tan, and Marie-Catherine de~Marneffe. 2023.
\newblock \href {https://doi.org/10.18653/v1/2023.findings-emnlp.712} {Ecologically valid explanations for label variation in {NLI}}.
\newblock In \emph{Findings of the Association for Computational Linguistics: EMNLP 2023}, pages 10622--10633, Singapore. Association for Computational Linguistics.

\bibitem[{Kim et~al.(2022)Kim, Lee, and Sohn}]{kim2022hate}
Jiyun Kim, Byounghan Lee, and Kyung-Ah Sohn. 2022.
\newblock Why is it hate speech? masked rationale prediction for explainable hate speech detection.
\newblock In \emph{Proceedings of the 29th International Conference on Computational Linguistics}, pages 6644--6655.

\bibitem[{Kurniawan et~al.(2025)Kurniawan, Mistica, Baldwin, and Lau}]{kurniawan2025training}
Kemal Kurniawan, Meladel Mistica, Timothy Baldwin, and Jey~Han Lau. 2025.
\newblock Training and evaluating with human label variation: An empirical study.
\newblock \emph{Computational Linguistics}, pages 1--27.

\bibitem[{Lin(2002)}]{lin2002divergence}
Jianhua Lin. 2002.
\newblock Divergence measures based on the shannon entropy.
\newblock \emph{IEEE Transactions on Information theory}, 37(1):145--151.

\bibitem[{Mathew et~al.(2021)Mathew, Saha, Yimam, Biemann, Goyal, and Mukherjee}]{mathew2021hatexplain}
Binny Mathew, Punyajoy Saha, Seid~Muhie Yimam, Chris Biemann, Pawan Goyal, and Animesh Mukherjee. 2021.
\newblock Hatexplain: A benchmark dataset for explainable hate speech detection.
\newblock In \emph{Proceedings of the AAAI conference on artificial intelligence}, volume~35, pages 14867--14875.

\bibitem[{Mendez~Guzman et~al.(2024)Mendez~Guzman, Schlegel, and Batista-Navarro}]{mendez2024outputs}
Erick Mendez~Guzman, Viktor Schlegel, and Riza Batista-Navarro. 2024.
\newblock From outputs to insights: a survey of rationalization approaches for explainable text classification.
\newblock \emph{Frontiers in Artificial Intelligence}, 7:1363531.

\bibitem[{Muscato et~al.(2025)Muscato, Passaro, Gezici, and Giannotti}]{muscato2025perspectives}
Benedetta Muscato, Lucia Passaro, Gizem Gezici, and Fosca Giannotti. 2025.
\newblock Perspectives in play: a multi-perspective approach for more inclusive nlp systems.
\newblock In \emph{Proceedings of the Thirty-Fourth International Joint Conference on Artificial Intelligence}, pages 9827--9835.

\bibitem[{Orgad et~al.(2026)Orgad, Barez, Haklay, Lee, Mosbach, Reusch, Saphra, Wallace, Wiegreffe, Wong, Tenney, and Geva}]{orgad2026actionable}
Hadas Orgad, Fazl Barez, Tal Haklay, Isabelle Lee, Marius Mosbach, Anja Reusch, Naomi Saphra, Byron~C. Wallace, Sarah Wiegreffe, Eric Wong, Ian Tenney, and Mor Geva. 2026.
\newblock \href {https://actionable-interpretability.github.io} {Interpretability can be actionable}.

\bibitem[{Plank(2022)}]{plank2022problem}
Barbara Plank. 2022.
\newblock The “problem” of human label variation: On ground truth in data, modeling and evaluation.
\newblock In \emph{Proceedings of the 2022 conference on empirical methods in natural language processing}, pages 10671--10682.

\bibitem[{Rizzi et~al.(2024)Rizzi, Leonardelli, Poesio, Uma, Pavlovic, Paun, Rosso, and Fersini}]{rizzi2024soft}
Giulia Rizzi, Elisa Leonardelli, Massimo Poesio, Alexandra Uma, Maja Pavlovic, Silviu Paun, Paolo Rosso, and Elisabetta Fersini. 2024.
\newblock Soft metrics for evaluation with disagreements: an assessment.
\newblock In \emph{Proceedings of the 3rd Workshop on Perspectivist Approaches to NLP (NLPerspectives)@ LREC-COLING 2024}, pages 84--94.

\bibitem[{Rodrigues and Pereira(2018)}]{rodrigues2018deep}
Filipe Rodrigues and Francisco Pereira. 2018.
\newblock Deep learning from crowds.
\newblock In \emph{Proceedings of the AAAI Conference on Artificial Intelligence}, volume~32. Association for the Advancement of Artificial Intelligence (AAAI).

\bibitem[{Sachdeva et~al.(2022)Sachdeva, Barreto, Bacon, Sahn, Von~Vacano, and Kennedy}]{sachdeva2022measuring}
Pratik Sachdeva, Renata Barreto, Geoff Bacon, Alexander Sahn, Claudia Von~Vacano, and Chris Kennedy. 2022.
\newblock The measuring hate speech corpus: Leveraging rasch measurement theory for data perspectivism.
\newblock In \emph{Proceedings of the 1st Workshop on Perspectivist Approaches to NLP@ LREC2022}, pages 83--94.

\bibitem[{Salles et~al.(2025)Salles, Vargas, and Benevenuto}]{salles2025hatebrxplain}
Isadora Salles, Francielle Vargas, and Fabr{\'\i}cio Benevenuto. 2025.
\newblock Hatebrxplain: A benchmark dataset with human-annotated rationales for explainable hate speech detection in brazilian portuguese.
\newblock In \emph{Proceedings of the 31st international conference on computational linguistics}, pages 6659--6669.

\bibitem[{Sandri et~al.(2023)Sandri, Leonardelli, Tonelli, and Je{\v{z}}ek}]{sandri2023don}
Marta Sandri, Elisa Leonardelli, Sara Tonelli, and Elisabetta Je{\v{z}}ek. 2023.
\newblock Why don’t you do it right? analysing annotators’ disagreement in subjective tasks.
\newblock In \emph{Proceedings of the 17th Conference of the European Chapter of the Association for Computational Linguistics}, pages 2428--2441.

\bibitem[{Sap et~al.(2020)Sap, Gabriel, Qin, Jurafsky, Smith, and Choi}]{sap2020social}
Maarten Sap, Saadia Gabriel, Lianhui Qin, Dan Jurafsky, Noah~A Smith, and Yejin Choi. 2020.
\newblock Social bias frames: Reasoning about social and power implications of language.
\newblock In \emph{Proceedings of the 58th annual meeting of the association for computational linguistics}, pages 5477--5490.

\bibitem[{Serdar et~al.(2021)Serdar, Cihan, Y{\"u}cel, and Serdar}]{serdar2021sample}
Ceyhan~Ceran Serdar, Murat Cihan, Do{\u{g}}an Y{\"u}cel, and Muhittin~A Serdar. 2021.
\newblock Sample size, power and effect size revisited: simplified and practical approaches in pre-clinical, clinical and laboratory studies.
\newblock \emph{Biochemia medica}, 31(1):27--53.

\bibitem[{Shapiro and Wilk(1965)}]{shapiro1965analysis}
Samuel~Sanford Shapiro and Martin~B. Wilk. 1965.
\newblock \href {https://doi.org/10.1093/biomet/52.3-4.591} {An analysis of variance test for normality (complete samples)}.
\newblock \emph{Biometrika}, 52(3-4):591--611.

\bibitem[{Sokolova and Lapalme(2009)}]{sokolova2009systematic}
Marina Sokolova and Guy Lapalme. 2009.
\newblock A systematic analysis of performance measures for classification tasks.
\newblock \emph{Information processing \& management}, 45(4):427--437.

\bibitem[{Strout et~al.(2019)Strout, Zhang, and Mooney}]{strout2019human}
Julia Strout, Ye~Zhang, and Raymond Mooney. 2019.
\newblock Do human rationales improve machine explanations?
\newblock In \emph{Proceedings of the 2019 ACL Workshop BlackboxNLP: Analyzing and Interpreting Neural Networks for NLP}, pages 56--62.

\bibitem[{Sullivan and Feinn(2012)}]{sullivan2012using}
Gail~M Sullivan and Richard Feinn. 2012.
\newblock Using effect size—or why the p value is not enough.
\newblock \emph{Journal of graduate medical education}, 4(3):279--282.

\bibitem[{Uma et~al.(2020)Uma, Fornaciari, Hovy, Paun, Plank, and Poesio}]{uma2020case}
Alexandra Uma, Tommaso Fornaciari, Dirk Hovy, Silviu Paun, Barbara Plank, and Massimo Poesio. 2020.
\newblock A case for soft loss functions.
\newblock In \emph{Proceedings of the AAAI Conference on Human Computation and Crowdsourcing}, volume~8, pages 173--177.

\bibitem[{Vargas et~al.(2022)Vargas, Carvalho, de~G{\'o}es, Pardo, and Benevenuto}]{vargas2022hatebr}
Francielle Vargas, Isabelle Carvalho, Fabiana~Rodrigues de~G{\'o}es, Thiago Pardo, and Fabr{\'\i}cio Benevenuto. 2022.
\newblock Hatebr: A large expert annotated corpus of brazilian instagram comments for offensive language and hate speech detection.
\newblock In \emph{Proceedings of the Thirteenth Language Resources and Evaluation Conference}, pages 7174--7183.

\bibitem[{Wilcoxon(1945)}]{wilcoxon1945individual}
Frank Wilcoxon. 1945.
\newblock Individual comparisons by ranking methods.
\newblock \emph{Biometrics bulletin}, 1(6):80--83.

\bibitem[{Zhao et~al.(2024)Zhao, Chen, Yang, Liu, Deng, Cai, Wang, Yin, and Du}]{zhao2024explainability}
Haiyan Zhao, Hanjie Chen, Fan Yang, Ninghao Liu, Huiqi Deng, Hengyi Cai, Shuaiqiang Wang, Dawei Yin, and Mengnan Du. 2024.
\newblock Explainability for large language models: A survey.
\newblock \emph{ACM Transactions on Intelligent Systems and Technology}, 15(2):1--38.

\bibitem[{Zhao and Aletras(2023)}]{zhao2023incorporating}
Zhixue Zhao and Nikolaos Aletras. 2023.
\newblock Incorporating attribution importance for improving faithfulness metrics.
\newblock In \emph{Proceedings of the 61st Annual Meeting of the Association for Computational Linguistics (Volume 1: Long Papers)}, pages 4732--4745.

\end{thebibliography}
